\documentclass[lettersize,journal]{IEEEtran}
\ifCLASSINFOpdf
\usepackage[pdftex]{graphicx}
\graphicspath{{../pdf/}{../jpeg/}}
\DeclareGraphicsExtensions{.pdf,.jpeg,.png}
\else
\usepackage[dvips]{graphicx}
\graphicspath{{../eps/}}
\DeclareGraphicsExtensions{.eps}
\fi

\usepackage[switch]{lineno}
\usepackage{ulem}
\usepackage{cancel}

\usepackage{epstopdf}
\usepackage{amsmath,amsthm, amssymb}
\usepackage{mathrsfs}
\usepackage{threeparttable}
\usepackage{bm}
\usepackage{multirow}
\usepackage{booktabs}
\usepackage{color}
\usepackage{tabularx}
\usepackage{setspace}
\usepackage{verbatim}
\usepackage{stfloats}
\usepackage{caption}
\usepackage{float}
\usepackage[T1]{fontenc}
\usepackage{mathrsfs}
\usepackage{soul}
\usepackage{cite}
\usepackage{color}
\usepackage{multirow}

\usepackage{stfloats}
\usepackage{mathtools}
\usepackage{amsmath}
\allowdisplaybreaks[0]
\usepackage{subfigure}

\usepackage{fancyhdr}
\usepackage{cancel}
\usepackage{array}

\usepackage{xcolor}
\makeatletter
\newif\if@restonecol
\makeatother

\usepackage[linesnumbered,ruled,vlined]{algorithm2e}
\usepackage{algpseudocode}
\usepackage{amsmath}
\usepackage{color}
\let\oldequation\equation
\let\oldendequation\endequation
\renewenvironment{equation}
{\linenomathNonumbers\oldequation}
{\oldendequation\endlinenomath}

\begin{document}
\normalem
		\title{DRL-Based Optimization for AoI and Energy Consumption in C-V2X Enabled IoV}
		
		\author{
			{
				Zheng Zhang, Qiong Wu, ~\IEEEmembership{Senior Member,~IEEE}, \\Pingyi Fan, ~\IEEEmembership{Senior Member,~IEEE}, Nan Cheng, ~\IEEEmembership{Senior Member,~IEEE}, \\Wen Chen, ~\IEEEmembership{Senior Member,~IEEE}, and  Khaled B. Letaief, ~\IEEEmembership{Fellow,~IEEE}
			}
			
			\thanks{
				{
				This work was supported in part by the National Natural Science Foundation of China under Grant 61701197; in part by the National Key Research and Development Program of China under Grant 2021YFA1000500(4); in part by Shanghai Kewei under Grant 22JC1404000 and 24DP1500500; in part by the Research Grants Council under the Areas of Excellence Scheme under Grant AoE/E-601/22-R; and in part by the 111 Project under Grant B23008. (Corresponding author: Qiong Wu)

				Zheng Zhang and Qiong Wu are with the School of Internet of Things Engineering, Jiangnan University, Wuxi 214122, China (e-mail: zhengzhang@stu.jiangnan.edu.cn, qiongwu@jiangnan.edu.cn).
							
				Pingyi Fan is with the Department of Electronic Engineering, State Key laboratory of Space Network and Communications, Tsinghua University, Beijing National Research Center for Information Science and Technology, Tsinghua University, Beijing 100084, China (email: fpy@tsinghua.edu.cn).
				
				Nan Cheng is with the State Key Lab. of ISN and School of Telecom-munications Engineering, Xidian University, Xi'an 710071, China (e-mail:
				dr.nan.cheng@ieee.org)
				
				Wen Chen is with the Department of Electronic Engineering, Shanghai JiaoTong University, Shanghai 200240, China (e-mail: wenchen@sjtu.edu.cn)
				
				K,B. Letaief is with the Department of Electrical and Computer Engineer-ing, the Hong Kong University of Science and Technology (HKUST), HongKong (email:eekhaled@ust.hk).
	
			}
}}
		
\maketitle
		
		\begin{abstract}
			To address communication latency issues, the Third Generation Partnership Project (3GPP) has defined Cellular-Vehicle to Everything (C-V2X) technology, which includes Vehicle-to-Vehicle (V2V) communication for direct vehicle-to-vehicle communication. However, this method requires vehicles to autonomously select communication resources based on the Semi-Persistent Scheduling (SPS) protocol, which may lead to collisions due to different vehicles sharing the same communication resources, thereby affecting communication effectiveness. Non-Orthogonal Multiple Access (NOMA) is considered a potential solution for handling large-scale vehicle communication, as it can enhance the Signal-to-Interference-plus-Noise Ratio (SINR) by employing Successive Interference Cancellation (SIC), thereby reducing the negative impact of communication collisions.
			When evaluating vehicle communication performance, traditional metrics such as reliability and transmission delay present certain contradictions. Introducing the new metric Age of Information (AoI) provides a more comprehensive evaluation of communication system. Additionally, to ensure service quality, user terminals need to possess high computational capabilities, which may lead to increased energy consumption, necessitating a trade-off between communication energy consumption and effectiveness.
			Given the complexity and dynamics of communication systems, Deep Reinforcement Learning (DRL) serves as an intelligent learning method capable of learning optimal strategies in dynamic environments. Therefore, this paper analyzes the effects of multi-priority queues and NOMA on AoI in the C-V2X vehicular communication system and proposes an energy consumption and AoI optimization method based on DRL. Finally, through comparative simulations with baseline methods, the proposed approach demonstrates its advances in terms of energy consumption and AoI.
		\end{abstract}

		
		\section{Introduction}
		\label{sec1}
		\IEEEPARstart{T}{he} Internet of Vehicles (IoV) technology is the foundation for the development of autonomous driving and intelligent transportation\cite{liu2021vehicular,wu2024cooperative}. In the future, IoV will be equipped with high-quality wireless services, including high-resolution radar perception capabilities and high-data-rate communication technologies between intelligent vehicles\cite{zhuang2019sdn,luo2023edgecooper,cui2024anti}. This will meet the diverse needs of users for vehicle applications such as automatic navigation, and collision warning, provided that timely access to the required data, videos, web pages, and other content is ensured for vehicle users through requests\cite{shen2023ringsfl,zhang2008network}. The traditional approach requires vehicles to first communicate with base stations,
		then accessing data stored in data centers through the access core backbone network, and finally, the data center returns the requested data\cite{wang2021cooperative,wang2010delay,fan2002investigation}.
		It is evident that this method experiences long end-to-end delays and has limited bandwidth for data return\cite{dai2019artificial,wu2023delay}.

		C-V2X is defined by 3GPP, with "V" representing vehicles and "X" indicating communication with other vehicles, network infrastructure, road infrastructure, pedestrians, and so on\cite{silva2017survey,shao2024semantic}. The issue of long end-to-end delays is addressed by V2V communication within this framework. In C-V2X mode 4, resources are autonomously selected by vehicular user equipment without reliance on cellular infrastructure, thus resolving issues of excessive delay and overhead\cite{molina2017lte,wang2024vehicle}. In mode 4, the automatic selection of communication resources by vehicles is termed as SPS, which is based on a distributed sensing-based semi-persistent resource scheduling protocol\cite{nabil2018performance}. Semi-persistence is exemplified by vehicles being able to choose one of the standardized Resource Reservation Intervals (RRI) and then determining the corresponding number of communication slots. However, this method may lead to collision risks as the same resources may be reserved by vehicles\cite{dayal2021adaptive,ji2024graph}.
		
		NOMA is a potential solution to address collision risks in C-V2X communication. It is a promising technology for handling large-scale vehicle communication \cite{Deng2023,yue2024hybrid}. NOMA can help alleviate performance degradation in terms of latency and packet reception probability caused by high vehicle density\cite{di2017non}. SIC is currently a detection technique can decode signals from multiple users occupying the same resource by leveraging their different signal strengths, thereby reducing interference in the SINR and minimizing latency, thus mitigating the impact of collisions\cite{zhang2017sic}.
		
		In communication systems, the environment can be influenced by various factors such as changes in user numbers, fluctuations of channel conditions, and the presence of interference\cite{wu2023characterizing,wu2022characterizing}. These factors make the optimal design of communication systems extremely complex\cite{wang2024value,wu2024urllc}. DRL algorithms can learn adaptive strategies in dynamic and complex environments\cite{Nan2019Space,shao2024semantic2}. Which can gradually learn the optimal strategy to achieve maximum long-term rewards, thus maximizing system performance\cite{leng2019agedeep,sun2024knowledge}. Which has been successfully applied in many fields, including but not limited to robotics, natural language processing, and vehicular communication networks\cite{zhang2024distributed,qi2024reconfigurable}.
		
		In existing engineering and 3GPP standards, AoI is the latest metric for measuring communication effectiveness\cite{leng2019AoI,qi2024deep}. This is because lower delay or higher reliability both result in lower average AoI, indicating better timeliness of communication\cite{yates2021age}. AoI refers to the time interval between the current time and the generation time of the data to be transmitted or received. If the average AoI at the receiving end is large, it means that the received data was generated a long time ago, indicating poor timeliness of communication. Currently, AoI is used in applications like ultra-reliable, industrial networks and low-latency communication.\cite{basnayaka2021age}. Therefore, low AoI facilitates the satisfaction of ultra-high data transmission rates, high-density connections, and high mobility to a great extent. However, to ensure these services, user terminals need to have high computing power for real-time signal processing, which may quickly deplete embedded batteries. Therefore, energy consumption of communication devices is also a concern\cite{hu2018integrated,qi2024reconfigurable2}.
		
		In summary, focusing on C-V2X vehicular networking, DRL, and performance optimization issues, considering the following factors such as the frequency of data generation from different vehicles, resource occupancy counts, resource occupancy collisions, communication energy consumption, and AoI is of great theoretical significance and practical value. The main contributions are as follows\footnote{
		{ The source code has been released at: https://github.com/qiongwu86/DRL-Based-Optimization-for-Information-of-Age-and-Energy-Consumption-in-C-V2X-Enabled-IoV
		}}:
		
		\begin{itemize}
			\item[1)] Consider the impact of message queuing in the C-V2X system on the AoI, along with the reconstruction of the vehicle's process of selecting resource blocks based on SPS, different arrival modes for various types of messages are modeled, and the AoI calculation model based on a multi-priority message queue involves tracking the AoI for messages with different priorities is established to analyze the role of multi-priority message queues in AoI.
			\item[2)]Examinate the impacts of message transmission in the C-V2X system on AoI, where utilizing NOMA technology at the receiver end with half-duplex resource selection schemes.It will give more details on the effects of NOMA in AoI.
			\item[3)] Propose a DRL scheme based on transmission power and interval in C-V2X Mode 4. Initially, vehicles reserve the resource blocks needed for the next period in the resource pool based on Mode 4 SPS selection, and obtain the number of time slots to be used according to the selected RRI. Subsequently, vehicles use broadcast communication to transmit higher priority messages to surrounding vehicles at the reserved resource's time slot. The receiver utilizes SIC-based NOMA technology to decode received messages separately. Finally, Roadside Units (RSUs) employ DRL to adjust the transmission intervals and transmit powers of each vehicle, finding the optimal strategy to optimize energy consumption and AoI in the C-V2X Mode 4 multi-type message scenario.
		\end{itemize}
		
		The rest of this paper is organized as follows. Section \ref{Related Work} reviews the related work. Section \ref{System Model And Problem Formulation} introduces the system model and formulates the optimization problem. Section \ref{DRL Method for Optimization of RRI and Power Allocation} simplifies the formulated optimization problem and presents the near optimal solution by DRL. We carry out some simulation to  demonstrate the effectiveness of our proposed method in Section \ref{Simulation Results and Analysis}, and conclude this paper in Section \ref{Conclusions}.
		
		\section{Related Work}
		\label{Related Work}
		With the increasing development of C-V2X, numerous research efforts have been initiated. In \cite{abbas2018novel}, a novel V2V resource allocation scheme using C-V2X technology is proposed to improve reliability and latency in connected vehicle networks. This scheme employs a hybrid architecture where V2V links are controlled by eNodeBs, with each vehicle periodically checking the lifespan of its data packets and requesting the eNodeB to allocate the best resources link to minimize overall latency.
		Li et al. in \cite{li2016analytical} utilized a Markov model combined with dynamic scheduling and SPS to assess available LTE idle resources for safety services. A weighted fair queuing algorithm is proposed to schedule safety beacons using LTE reserved resources, improving safety application reliability within limited LTE bandwidth.
		Authors in \cite{molina2017system} pointed out that there are four types of transmission errors in C-V2X mode 4, including half-duplex errors, resource duplication, insufficient link budget, and failures due to channel fading. The first two are caused by the unique SPS resource selection scheme at the MAC layer.
		
		NOMA, based on power-domain multiplexing, can enhance spectrum efficiency by allowing different users to share the same resources with power multiplexing. In \cite{xu2020energy}, the authors addressed the random access problem in NOMA-based backscatter communication networks with quality of service guarantees for the first time. They developed an iterative algorithm using the Dinkelbach method and quadratic transformation method to maximize user energy efficiency under constraints such as meeting QoS requirements and continuous interference elimination for SIC.
		\cite{zhang2020energy} investigated energy efficiency maximization in NOMA-MIMO systems based on terahertz (THz) communication for the first time. They proposed a rapid convergence scheme for user clustering in THz-NOMA-MIMO systems using an enhanced K-means learning algorithm based on channel correlation characteristics. This approach maximizes the energy efficiency of cache-enabled systems even with imperfect SIC, achieving faster convergence and lower power consumption, thus realizing higher energy efficiency in terahertz cache networks.
		Seo et al. in \cite{seo2018nonorthogonal} studied NOMA random access based on channel inversion with multi-level target powers. This ensures that the receiving power at the base station can be one of two target values to guarantee the power difference for SIC. Compared to OMA, the maximum receiving power can be increased from 0.368 to 0.7. NOMA shows potential in optimizing communication performance by improving energy efficiency.
		
		In recent years, DRL has become commonplace for performance optimization. Ron et al. proposed a method for optimizing device transmission power to mitigate interference in \cite{ron2021drl}. They pointed out that traditional power allocation problems are often modeled as non-deterministic polynomial time hard (NP-hard) combinatorial optimization problems with linear constraints, making traditional optimization methods ineffective. Therefore, they employed DRL algorithms to optimize vehicle transmission power and achieve D2D communication under cellular network conditions, demonstrating the potential advantages and effectiveness of DRL algorithms in optimizing power in communication.
		In \cite{wang2021joint}, a novel approach combining resource allocation, power control, and DRL was proposed to improve D2D communication quality in critical mission communications. The DRL method based on spectrum allocation strategy enables D2D users to autonomously select channels and power, significantly improving system capacity and spectrum efficiency while minimizing interference to cellular users. Experimental results demonstrated that the method effectively improves resource allocation and power control, providing an efficient optimization solution for critical mission communications.
		In \cite{ciftler2021distributed}, the authors explored resource allocation problems in hybrid radio-frequency and visible light communication networks to improve network throughput and energy efficiency. Traditional methods perform poorly when faced with dynamic channels and limited resources, and heuristic methods have limited adaptability to channel and user demand changes, requiring increased communication overhead. Therefore, distributed algorithm based on DRL can adapt to network changes and optimize user transmission power to achieve target data rates, providing an effective solution for resource allocation problems in hybrid radio-frequency and visible light communication networks. This is the motivation of us doing this work. 
		
		The Information Age is used to analyze applications with timeliness requirements, mainly measure the timeliness of information in queues and transmission systems. Literature \cite{yates2012real} analyzed the average AoI in multi-source node queuing systems considering data packet Poisson arrivals and exponential service times under Last-Come First-Served (LCFS) and First-Come First-Served (FCFS) queuing disciplines. A new, lower complexity simplification technique was proposed to evaluate the average AoI in finite-state continuous-time queuing systems. \cite{he2017optimal} considered scenarios where multiple source node-destination node pairs share the same channel and tried to minimize the average age of all source node-destination node pairs through link scheduling. It was first proven that the age minimization link scheduling problem is an NP-hard problem. Then, an optimal solution was given using integer linear programming methods, and a highly scalable steepest age descent algorithm was proposed, which has near-optimal performance and fast convergence speed.
		
		Based on our investigation, there is no existing work in the current C-V2X vehicular networking environment that considers the role of multi-priority message queues in various aspects of message AoI, and optimizes communication performance by mitigating the impact of resource collisions through NOMA technology in this scenario. Which motivates us to conduct this work. Furthermore, building upon this foundation, we aim to utilize receiver AoI and communication energy consumption as performance metrics for the V2V direct communication mode of C-V2X. Additionally, we will further consider the resource selection characteristics of C-V2X and employ a DRL algorithm for RRI and vehicle transmission power, while ensuring lower AoI and communication energy consumption.
		
		\section{System Model And Problem Formulation}
		\label{System Model And Problem Formulation}
		\begin{figure*}
			\centering
			\includegraphics[width=0.8\textwidth,trim=2bp 1bp 4bp 3bp, clip]{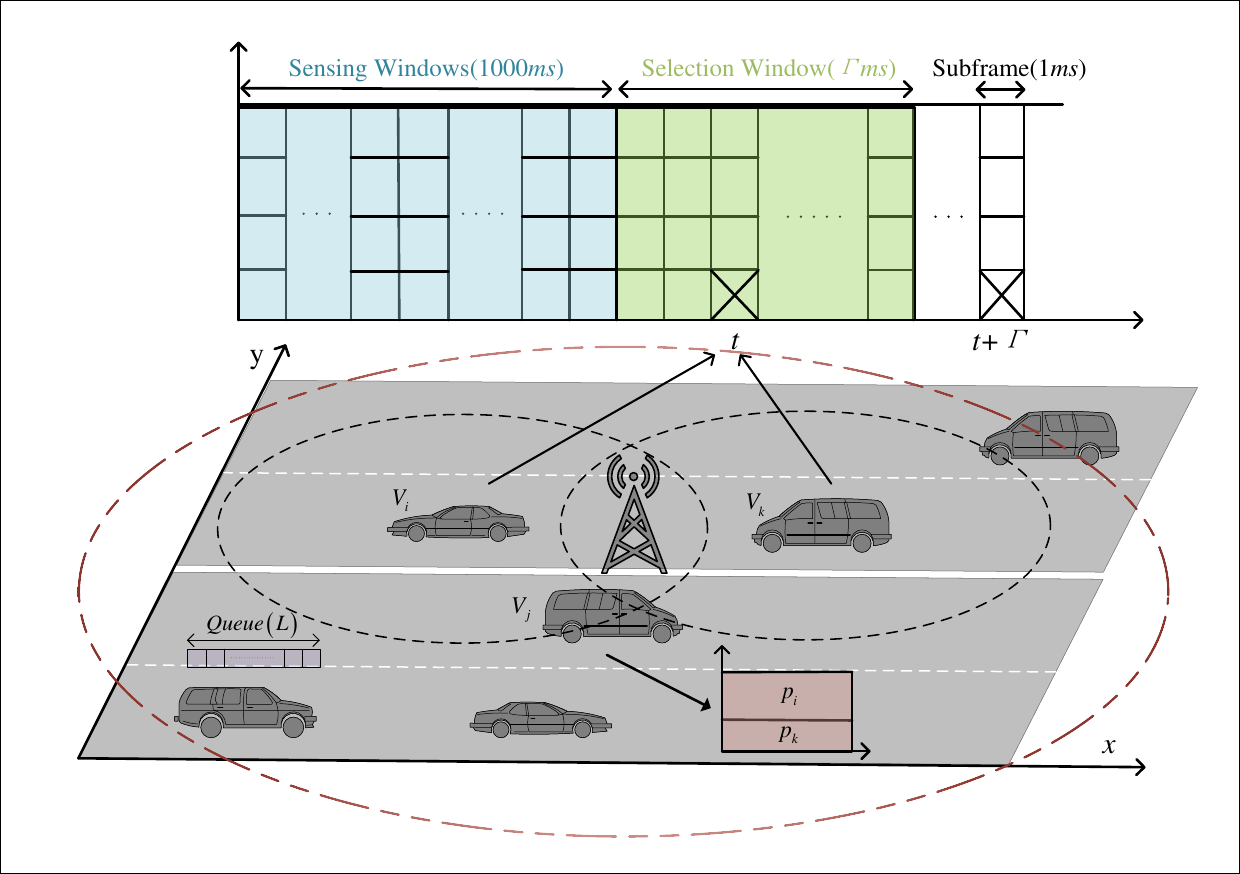}
			\caption{C-V2X IoV system}
			\label{fig1}
			\vspace{-0.5cm}
		\end{figure*}
		
		\begin{table*}
			\caption{The summary for notations.}
			\label{tab1}
			\centering
			\setlength{\tabcolsep}{0.5mm}{
			\begin{tabular}{|c|p{6.5cm}|c|p{6.5cm}|}
				\hline
				\textbf{Notation} &\textbf{Description} &\textbf{Notation} &\textbf{Description}\\
				\hline
				
				{$N_v$} &\multicolumn{1}{m{6.5cm}|}{The total count of vehicles.} 
				&$D$ &\multicolumn{1}{m{6.5cm}|}{The length of the highway.}\\
				\hline
				
				{$y^t_i$} &\multicolumn{1}{m{6.5cm}|}{The distance between vehicle i along the y-axis and the origin at time slot $t$.} 
				&$x^t_i$ &\multicolumn{1}{m{6.5cm}|}{The distance between vehicle i along the x-axis and the origin at time slot $t$.}\\
				\hline
				
				$v$ &Vehicle speed. &$F$ &\multicolumn{1}{m{6.5cm}|}{Number of lanes.}\\
				\hline
				
				$f$ &Lane index. &$d_y$ &\multicolumn{1}{m{6.5cm}|}{The width of a lane.}\\
				\hline
				
				$w$ &\multicolumn{1}{m{6.5cm}|}{Maximum distance at the receiver.}
				&$L$ &\multicolumn{1}{m{6.5cm}|}{The maximum length of the message queue.}\\
				\hline
				
				$RB^t_i$ &\multicolumn{1}{m{6.5cm}|}{The time slot where vehicle $i$ reserves resources. }
				&$\gamma$ &\multicolumn{1}{m{6.5cm}|}{Discount factor.}\\
				\hline
				
				$t_{rk}$ &\multicolumn{1}{m{6.5cm}|}{The time slot when the vehicle is preparing to re-reserve resources.} 
				&{$m^t_i$} &\multicolumn{1}{m{6.5cm}|}{The remaining amount of reserved resources that can be used.} \\
				\hline
				
				$T_w$ &\multicolumn{1}{m{6.5cm}|}{Resource selection buffer time.}
				&$T_s$ &Select the time slot to reserve resources. \\
				\hline
						
				$SPS$ &\multicolumn{1}{m{6.5cm}|}{The method for vehicles to autonomously choose communication resources. }
				&$SW$ &\multicolumn{1}{m{6.5cm}|}{The value of the reselection counter for vehicles.}\\
				\hline
				
				$RC$ &\multicolumn{1}{m{6.5cm}|}{The value of the reselection counter for vehicles.}
				&$RRI$ &\multicolumn{1}{m{6.5cm}|}{Resource selection buffer time.}\\
				\hline
				
				$\Gamma^t_i$ &\multicolumn{1}{m{6.5cm}|}{The RRI size selected by vehicle $i$.}
				&$\beta^t_i$ &Has vehicle $i$ transmitted messages.\\
				\hline
				
				$\alpha _i^t$ &Is there a message generated for vehicle $i$. 
				&${K_H}$ &\multicolumn{1}{m{6.5cm}|}{Number of repeated transmissions of HPD.}\\
				\hline
				
				$K_D$ &{Number of repeated transmissions of DENM.} 
				&{$T_H$} &\multicolumn{1}{m{6.5cm}|}{{HPD repeat transmission interval.}}\\
				\hline
				
				{$T_D$} &\multicolumn{1}{m{6.5cm}|}{DENM repeat transmission interval.}
				&$\rho $ &\multicolumn{1}{m{6.5cm}|}{Whether each priority message was transmitted.}\\
				\hline
				
				$b$  &\multicolumn{1}{m{6.5cm}|}{The position of the message in queue. }
				&$\varphi _{i,n}^{t,b}$ &\multicolumn{1}{m{6.5cm}|}{AoI of messages at position $b$ in queue $n$ of $i$.}\\
				\hline
				
				$\Phi _{i \to j}^t$ &the AoI of the receiving end $j$ for vehicle $i$. 
				&$u^t_i$ &Indicates that $i$ transmitted the message to $j$.\\
				\hline
				
				$W$ &Communication resource channel bandwidth. 
				&$R^t_{th}$ &Transmission rate threshold.\\
				\hline
				
				$G$ &The size of the transmitted message. 
				&$\eta $ &SINR for communication.\\
				\hline
				
				$p_i^t$ &\multicolumn{1}{m{6.5cm}|}{The transmission power of vehicle $i$ at time slot $t$.} 
				&$h$ &\multicolumn{1}{m{6.5cm}|}{Channel gain.}\\
				\hline
				
				$p_n^2$ &\multicolumn{1}{m{6.5cm}|}{Noise energy.} 
				&$l_i$ &\multicolumn{1}{m{6.5cm}|}{Communication time of vehicle $i$.}\\
				\hline
				
				$P_{col}$ &\multicolumn{1}{m{6.5cm}|}{Resource collision probability.} 
				&$\pi$ &\multicolumn{1}{m{6.5cm}|}{The probability that the vehicle is at the moment when it is ready to select resources.}\\
				\hline
				
				$CSR$ &\multicolumn{1}{m{6.5cm}|}{The number of resources in SW.} &$\varepsilon _i^t$ &\multicolumn{1}{m{6.5cm}|}{Energy consumption of vehicle $i$ in time slot $t$.}\\
				\hline
				
				$E^t_i$ &\multicolumn{1}{m{6.5cm}|}{The total energy consumption generated by vehicle $i$ occupying a reserved resource.} 
				&$\bar E$ &Average energy consumption.\\
				\hline
				
				$\bar \Phi $ &Average AoI in receiver. 
				&${\omega _1}$ &\multicolumn{1}{m{6.5cm}|}{The weighting factor of energy consumption in the reward function.}\\
				\hline
				
				${\omega _2}$ &\multicolumn{1}{m{6.5cm}|}{The weight factor of AoI in the reward function.} &$a^t_i$ &\multicolumn{1}{m{6.5cm}|}{The action of vehicle $i$ in time slot $t$.}\\
				\hline
				
				$s^t_i$ &The state of vehicle $i$ in time slot $t$. 
				&$N_i$ &Number of receivers for vehicle $i$.\\
				\hline
				
				$d_i$ &\multicolumn{1}{m{6.5cm}|}{The average distance between vehicle i and the receiver.} 
				&$Rn$ &\multicolumn{1}{m{6.5cm}|}{The ratio of receivers who successful communications to all receivers.}\\
				\hline
				
				$\theta_Q$ &\multicolumn{1}{m{6.5cm}|}{The weights of the state-action value function in strategy networks. }
				&$\theta_x$ &\multicolumn{1}{m{6.5cm}|}{The weights of the transition function strategy network.}\\
				\hline
				
				$lr_Q$ &\multicolumn{1}{m{6.5cm}|}{Learning rate of value Q approximation function. }
				&$lr_x$ &\multicolumn{1}{m{6.5cm}|}{Learning rate of state action transition approximation function.}\\
				\hline
				
				$\tau_Q$ &\multicolumn{1}{m{6.5cm}|}{The objective change rate of the approximate function of the value function. }
				&$\tau_x$ &\multicolumn{1}{m{6.5cm}|}{The objective change rate of the approximate function of state action transition.}\\
				\hline
				
				$M$ &Replay memory size. 
				&$B$ &Sample size.\\
				\hline
			\end{tabular}
		}
			\vspace{-0.5cm}
		\end{table*}
		
		\subsection{Scenario description}
		Fig. \ref{fig1} shows a C-V2X vehicular networking system. Assuming a highway divided into two lanes in each direction, with vehicles randomly distributed and moving at uniform speeds. Vehicles closer to the center lane travel faster, mimicking real highway scenarios. Each vehicle considers other vehicles within a circular area of radius $w$ as potential receivers. Each vehicle maintains a message queue to store newly generated but unprocessed message data, with a maximum queue length of $L$, following the policy of FCFS. If the number of messages in the queue reaches $L$, newly generated messages are discarded due to overflow.
		
		Each vehicle in C-V2X utilizes the SPS scheme to reserve communication resources for itself. When new messages enter the message queue, vehicles check if they have any reserved communication resources. If not, they establish a Selection Window (SW) at that moment. The SW serves as a time window, indicating the duration within which the vehicle can reserve resources for itself in the future. This time window, along with its corresponding frequency domain, forms a time-frequency multiplexing structure capable of providing communication resources, known as the resource pool. Vehicles, after partitioning the resource pool, assess the occupancy of resources in the previous one thousand subframes based on metrics like Reference Signal Received Power (RSRP) and Received Signal Strength Indicator (RSSI) to determine suitable resources for communication. They then randomly reserve some of these resources for future use. Vehicles determine the RRI based on the size of the SW, i.e., the duration of the resource pool. Additionally, vehicles calculate the number of times they can use the resources according to the protocol and store it as a Reselection Counter (RC). Each time the resources are used for transmission, the RC is decremented until it reaches zero, at which point the vehicle waits for the process to repeat when a new message enters the queue. This resource selection method can lead to overlapping SW when multiple vehicles determine their SW at roughly similar times, resulting in collisions where they reserve the same resource for future broadcast communication.
		
		When many vehicles simultaneously occupy the same resource, then the vehicles as receivers use SIC-based NOMA to decode messages in descending order of power. Lower-power messages are treated as interference and excluded after decoding the highest power message, thereby improving the SINR of each message and reducing the AoI resulting from communication failures. AoI reflects the time spent in queuing and transmission processes, where the queuing process is affected by packet generation rate or RRI, and the other process is influenced by success rate or delay. Reducing RRI decreases AoI but increases energy consumption and collision probability, affecting SINR and increasing AoI. Reducing transmission power can decrease energy consumption but also reduces SINR, leading to increased AoI. Hence, there is a need to mitigate collision effects and jointly optimize energy consumption and AoI.
		
		An RSU with a coverage radius of $R$ is placed at the roadside, capable of sending resource selection notifications to vehicles entering its coverage area. In each time slot, vehicles check whether resource reselection is needed and select RRI size and transmission power based on information provided by the RSU. Through trained policies, the RSU can choose appropriate actions based on vehicle status to optimize AoI and energy consumption.
		Table \ref{tab1} summarizes the symbols used in this part.
		
		\subsection{Vehicle motion model}
		Assuming all vehicles are randomly distributed across various lanes of a bidirectional highway, let $(x, y)$ denote the position of vehicle $i$ at time $t$, with the lower-left corner of the highway in Fig. \ref{fig1} as the coordinate origin. Let $x$ and $ y$ represent the distances of vehicle $i$ along the x-axis and y-axis from the origin at time $t$, respectively. It is assumed that the position of each vehicle is updated with each time step
		\begin{equation}
			\begin{aligned}
				x_i^{t + 1} = x_i^t + \delta v_i^f\tau ,x_i^t \in [0,D]
			\end{aligned}  ,
			\label{x}
		\end{equation}
		where $D$ represents the length of the highway, $\delta$ indicates the direction of vehicle travel, where $\delta=1$ represents the x-axis direction, and $\delta=-1$ represents another direction. 
		The speed $v^f_i$ is defined as 
		${v_{max}} - 20\left| {{f_i} - {\raise0.7ex\hbox{$F$} \!\mathord{\left/
					{\vphantom {F 2}}\right.\kern-\nulldelimiterspace}
				\!\lower0.7ex\hbox{$2$}} + {\raise0.7ex\hbox{${\left( {\delta  - 1} \right)}$} \!\mathord{\left/
					{\vphantom {{\left( {\delta  - 1} \right)} 2}}\right.\kern-\nulldelimiterspace}
				\!\lower0.7ex\hbox{$2$}}} \right|$
		on lane $f$, with $v_{max}$ representing the maximum vehicle speed. The relationship between $y$ and the lane index $f$ of vehicle $i$ is as follows
		\begin{equation}
			\begin{aligned}
				y_i^{} = {f_i}d_y^{} - {y_0}
			\end{aligned}  ,
			\label{y}
		\end{equation}
		where $d_y$ represents the distance between two lanes in the y-axis direction on the highway, and $y_0$ denotes half of $d_y$.

		\subsection{Resource reservation model}
		In C-V2X Mode 4, the process of resource reservation using SPS is mainly divided into three steps: (1) When vehicles transmit a new packet and the RC is 0, it must reserve new resources within the SW. (2) The vehicle first creates a resource list, $L_A$, which contains resources that can be reserved. These resources include those that may be occupied by other vehicles within a certain period after the packet is generated, provided that the RSRP exceeds a threshold, or resources that have been occupied by the vehicle itself in each SW-sized time interval before the packet generation. (3) From $L_A$, the vehicle selects $20\%$ of the total number of resources with the lowest RSSI to create list $L_C$, for random reservation by the vehicle.
		
\begin{figure}[h]
			\centering
			\includegraphics[width=0.45\textwidth,trim=5bp 2bp 2bp 1bp, clip]{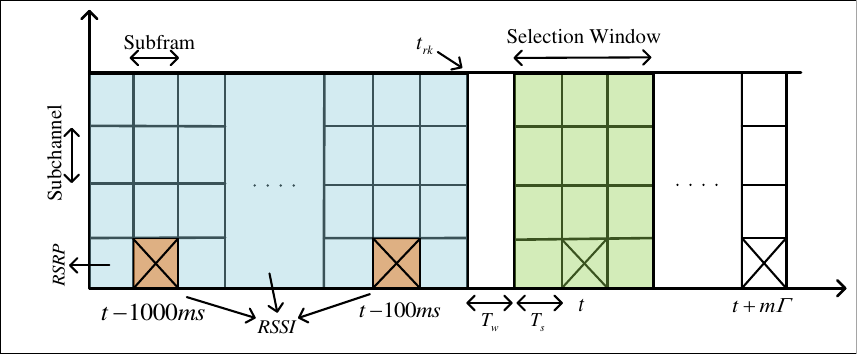}
			\caption{Schematic diagram of communication resource reservation}
			\label{fig2}
			\vspace{-0.5cm}
		\end{figure}

		Due to the periodic nature of resource occupation by vehicles using SPS, assumed to be the time slots where reserved resources are located. When the RC of the currently occupied resources becomes 0, it needs to be determined whether the queue is empty. If it is not empty, there is a probability $P_{rk}$ of continuing to use the currently occupied resources, and a probability of $1-P_{rk}$ of re-selecting resources. In both cases, RC is reassigned, the time slot of the resources that the vehicle will occupy is represented as
		\begin{equation}
			\begin{aligned}
				RB_i^t = {t_{rk}} + \Gamma _i^t + m_i^t\Gamma _i^t
			\end{aligned}  ,
			\label{RB1}
		\end{equation}
		\begin{equation}
			\begin{aligned}
				RB_i^t = {t_{rk}} + {T_w} + {T_s} + \Gamma _i^t + m_i^t\Gamma _i^t
			\end{aligned}  ,
			\label{RB2}
		\end{equation}
		the variables $t_{rk}$, $\Gamma$, $m$, $T_w$, and $T_s$ represent the time slot for vehicles to prepare for re-reserving resources, the size of RRI, the remaining number of times the current resources can be occupied (RC does not decrease if no packets are sent during occupancy), the resource selection buffer time from $0$ to $3$, and the time interval from the start of the vehicle's SW to the subframe where the resource is selected, as shown in Fig. \ref{fig2}. 
		And the $t$ and $t+m\Gamma$ in Fig. \ref{fig2} represent the position (space occupying) of $RB_i^t$. Unlike Eq.\ref{RB2}, the situation represented by Eq.\ref{RB1} does not require waiting for $T_w$ and $T_s$.
		
		If the queue is determined to have no messages in queue, the vehicle needs to wait for the arrival of the first message in the queue before starting to select resource blocks, then RB is represented as
		\begin{equation}
			\begin{aligned}
				RB_i^t = {t_a} + {T_w} + {T_s} + {\Gamma _t} + m_i^t{\Gamma _t}
			\end{aligned}  ,
			\label{RB3}
		\end{equation}
		where $t_{a}$ is the arrival time of the first packet in the queue. When $RC \neq 0$, its expression along with $m$ is as follows:
		\begin{equation}
			\begin{aligned}
				RC_i^{t + 1} = RC_i^{\rm{t}} - \beta _i^t
			\end{aligned}  ,
			\label{RC}
		\end{equation}
		\begin{equation}
			\begin{aligned}
				m_i^{t + 1} = m_i^t + \left\lfloor {\frac{t}{{RB_i^t}}} \right\rfloor \left\lceil {\frac{{RC_i^t}}{{RC_i^0}}} \right\rceil 
			\end{aligned}  ,
			\label{m}
		\end{equation}
		where $\beta$ is a binary variable indicating whether the vehicle transmits messages from the queue, and $RC^{0}_{i}$ is the initial value of RC.
		
		\subsection{Message queue model}
		The status of the message queue depends on message generation and processing. Assuming the queue has a capacity of $L$, the length of the queue, $q$, can be be updated as follows
		\begin{equation}
			\begin{aligned}
				q_i^{t + 1} = \left\{ {\begin{array}{*{20}{c}}
						{q_i^t}&{\alpha _i^t = 0,\beta _i^t = 0}\\
						{q_i^t - 1}&{\alpha _i^t = 0,\beta _i^t = 1}\\
						{q_i^t + 1}&{\alpha _i^t = 1,\beta _i^t = 0}\\
						{q_i^t}&{\alpha _i^t = 1,\beta _i^t = 1}\\
						L&{q_i^t = L,\beta _i^t = 0}
				\end{array}} \right.
			\end{aligned}  ,
			\label{q}
		\end{equation}
		where $\alpha^{t}_{i}$ represents whether vehicle $i$ generates a message at time slot $t$. This queue applies to all message types in C-V2X.
		
		C-V2X mode 4 includes four types of messages: High-Priority Data (HPD), Decentralized Environmental Notification Messages (DENM), Cooperative Awareness Messages (CAM), and Miscellaneous High-Density Data (MHD). Their priority order is HPD > DENM > CAM > MHD. CAM-type packets are generated periodically with a period of $\frac{1}{T_c}$, while other types of packets are generated trigger-based. Therefore, the expression for whether CAM messages are generated is
		\begin{equation}
			\begin{aligned}
				\alpha _{i,c}^t = 1 - \left\lceil {\frac{{t\bmod {T_c}}}{{{T_c}}}} \right\rceil 
			\end{aligned}  ,
			\label{a_c}
		\end{equation}
		when $t$ is an integer multiple of $T_c$, $a^{t}_{i} = 1$, indicating the occurrence of a new CAM message. 
		The probability of new packet generation for other types follows the probability mass function of the Poisson distribution
		\begin{equation}
			\begin{aligned}
				P(arr_{i,n}^t = 1) = {\lambda _n}{e^{ - {\lambda _n}}}
			\end{aligned}  ,
			\label{a_n}
		\end{equation}
		where $arr_{i,n}^t = 1$ indicates that a new package has been generated, $\lambda_n$ is the arrival rate, $n$ encompasses the production quantities of the three message types mentioned above within a certain time period. To ensure the successful transmission of new messages generated by HDM and DENM, each needs to be retransmitted $K_{H}$ and $K_{D}$ times respectively, with this retransmission process occurring periodically at intervals of $T_{H}$ and $T_{D}$. At this point, the queue arrival rate is the sum of the production rates of the four message types, while the processing rate is the reciprocal of the RRI, $\frac{1}{\Gamma}$.
		
		In a single priority queue, $\beta_i^t$ is equivalent to $\rho_i^t$, representing the successful transmission of different message types in the single priority queue. This occurs when there are messages queued in the vehicle queue and reserved communication resources are available for use at that moment. The expression is as follows
		\begin{equation}
			\begin{aligned}
				\rho _i^t = \left\lfloor {\frac{t}{{RB_i^t}}} \right\rfloor \left\lceil {\frac{{q_i^t}}{L}} \right\rceil 
			\end{aligned}  .
		\end{equation}
		
		The expression indicates that the transmission operation can only be completed when the estimated time equals the current time and the queue is not empty.
		
		When operating under a single priority scheme, multiple message types share the total message generation rate and processing rate. However, in a multi-priority scheme, each message type has its own message generation rate, and the processing rate for high-priority messages equals the total processing rate of the single-priority queue. Therefore, in a multi-priority queue, the ratio of the message generation rate to the processing rate for high-priority messages is relatively low, ensuring a lower AoI for high-priority messages.
		
		Therefore, four corresponding FIFO queues are established for different types of signals, all with a capacity of $L$ and a queue length of $q$. New messages of the corresponding type can only be added to the queue when the queue satisfies $q < L$, and transmission opportunities leaving the queue need to prioritize high-priority messages. Transmission opportunity refers to the vehicle being able to use reserved resources in a particular time slot $t$. Therefore, the expressions for the transmission action $\beta$ and the determination of whether each queue can be transmitted $\rho$ in the multi-priority queue are as follows
		\begin{equation}
			\begin{aligned}
				\left\{ {\begin{array}{*{20}{c}}
						{\beta _{i,H}^t = 1,\beta _{i,C}^t = 0}&{\rho _{i,H}^t = 1}\\
						{\beta _{i,H}^t = 0,\beta _{i,C}^t = 1}&{\rho _{i,C}^t(1 - \rho _{i,H}^t) = 1}\\
						{\beta _{i,H}^t = \beta _{i,C}^t = 0}&{otherwise}
				\end{array}} \right.
			\end{aligned}  ,
		\end{equation}
		
		\begin{equation}
			\begin{aligned}
				\rho _{i,n}^t = \left\lfloor {\frac{t}{{RB_i^t}}} \right\rfloor \left\lceil {\frac{{q_{i,n}^t}}{L}} \right\rceil 
			\end{aligned}  ,
		\end{equation}
		the $n$ in $\rho^{t}_{i,n}$ includes four types of messages, such as $\rho^{t}_{i,H}$ and $\rho^{t}_{i,C}$. 
		Because it is determined in order of priority, the expression involving two priority queues can reflect the relationship between different queue $\beta$ values.

		\subsection{AoI model}
		\begin{figure}[h]
			\centering\
			\includegraphics[width=0.45\textwidth,trim=60bp 2bp 45bp 2bp, clip]{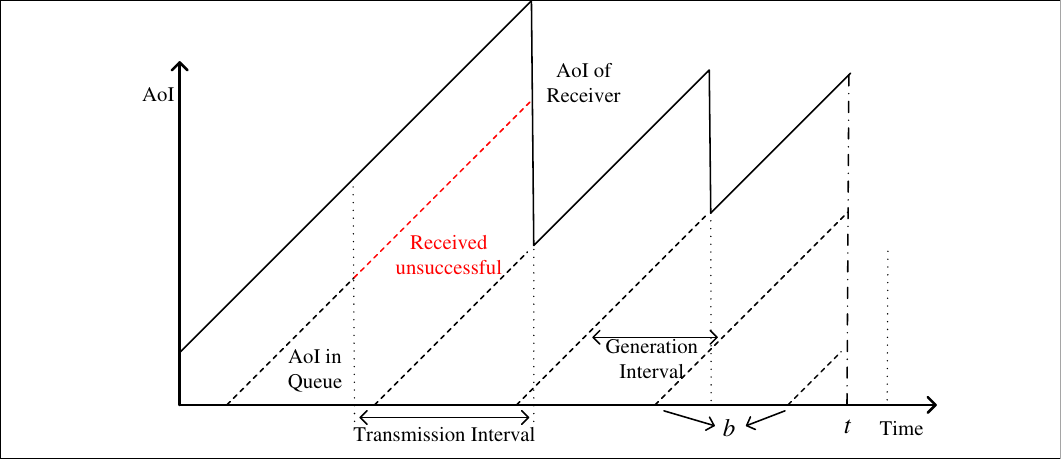}
			\caption{Changes in AoI.}
			\label{AoImodel}
			\vspace{-0.5cm}
		\end{figure}
		
		The changes in AoI are shown in Fig. \ref{AoImodel}. The solid line represents the AoI of receiver , the dark dashed line represents the AoI of the messages in the queue, both of which increase over time. The light dashed line represents the time slot that the vehicle can occupy for resource transmission. If the transmission is successful, the AoI of receiver will be updated to the AoI of received message. The red dashed line indicates that the message transmission has failed, and the receiver is unable to update the AoI, so it will continue to grow on the previous basis. In addition, $b$ represents the position of each message in the queue.
		
		The AoI increases with each subframe while messages are queued, but it stops increasing when they leave the queue. Therefore, the AoI expression for four types of packets in corresponding queue is
		\begin{equation}
			\begin{aligned}
				\varphi _{i,n}^{t + 1,b} = \left\{ {\begin{array}{*{20}{c}}
						{\varphi _{i,n}^{t,b + 1} + 1}&{\beta _{i,n}^t = 1}\\
						{\varphi _{i,n}^{t,b} + 1}&{\beta _{i,n}^t = 0}
				\end{array}} \right. 
			\end{aligned}  ,
		\end{equation}
		where $n$ encompasses four queues, $\phi^{t,b}_{i,n}$ represents the AoI of the message at position $b$ in queue $n$ of vehicle $i$ at time slot $t$. When $\beta$=1, the positions of all packets in the queue, except for the head, will shift. The probability of $\beta$=1 is the processing rate, since the processing rate in C-V2X is expressed as $\frac{1}{\Gamma}$, and the RRI will affect AoI of the message in queue.
		
		The accumulated AoI in communication process represents the time elapsed between two consecutive transmissions from the transmitter to the receiver. 
		Moreover, since received messages reflect the transmitter's conditions at transmission time, the receiver must retain their AoI to show the freshness of the data about the transmitter.
		The expression for the AoI of vehicle $j$ as receiver with respect to vehicle $i$ as transmitter  
		\begin{equation}
			\begin{aligned}
				\Phi_{i \rightarrow j}^{t+1}=\left\{\begin{array}{cc}
					\varphi_{i n}^{t .1}+l_{i \rightarrow j}^{t} & u_{i \rightarrow j}^{t}=1 \\
					\Phi_{i \rightarrow j}^{t}+1 & u_{i \rightarrow j}^{t}=0
				\end{array}\right.
			\end{aligned}  ,
		\end{equation}
		where $\Phi^t_{i \rightarrow j}$ represents the AoI of vehicle $j$ with respect to vehicle $i$ at time $t$, $l$ denotes the transmission delay, and $u$ represents the transmission situation. When $u=1$, it indicates that vehicle $i$ successfully transmits the message to $j$. In this case, the AoI at vehicle $j$ is calculated by adding the transmission time to the AoI of the packet at the head of the highest priority queue broadcasted by vehicle $i$. When $u=0$, it signifies a transmission failure, and thus $\Phi^t_{i \rightarrow j}$ is updated to $\Phi^{t-1}_{i \rightarrow j}$ plus one time slot. Due to the unique way of allocating resources for vehicles in C-V2X mode 4, each transmission failure necessitates waiting for a time interval equal to the RRI size, resulting in an increase in $\Phi^t_{i \rightarrow j}$ by $\Gamma$.

		\subsection{Communication model }
		The size of the resources in mode 4 is variable, and the vehicle occupies the required bandwidth \(W\) based on the message size $G$, while the expression for the transmission rate threshold is
		\begin{equation}
			\begin{aligned}
				R_{th}^t = W_i^{}{\log _2}(1 + {\eta _{th}})
			\end{aligned}  ,
		\end{equation}
		
		Among them $R_{\text{th}} = G$, because the time for the data packet to complete data transmission should not exceed one slot, that is, $\frac{G}{R} \leq 1$. Then calculate the SINR threshold based on the bit width and $R_{th}^t$ as follows
		\begin{equation}
			\begin{aligned}
				{\eta _{th}} = {2^{R_{th}^t/{B_i}}} - 1 = {2^{{G_i}/{B_i}}} - 1
			\end{aligned}  ,
		\end{equation}
		
		Different distances between vehicles at different times cause varying channel damage in communication, resulting in the receiving end receiving at different communication rates on different channels. And vehicles can use different resources at the same time, so the vehicle as receiver gets 
		message at varying rates across these resources. When the vehicle only receives one 
		message at most within a resource, at time slot $t$ the SINR is
		\begin{equation}
			\begin{aligned}
				\eta _{i \to r}^{t,n} = \frac{{p_{i}^t{{\left| {h_{i \to j}^{t,n}} \right|}^2}}}{{p_n^2}}
			\end{aligned}  ,
		\end{equation}
		$p^t_i$ is the transmission power of vehicle $i$, $\lvert h^{t,n}_{i \rightarrow j} \lvert^2$ is the channel gain for communication, $p_n$ is the noise energy, and $n$ is the communication resource.
		
		However, vehicles in the C-V2X model 4 system may reserve the same resource, and the collisions may occur. According to \cite{nba2020discrete}, the collision probability in the system can be expressed as
		\begin{equation}
			\begin{aligned}
				\begin{array}{l}
					P_{col}^{} \approx 
					1 - {\left[ {1 - \left[ {1 - \prod\limits_{{\rm{i}} = 0}^{\Gamma  - 1} {(1 - \frac{\pi }{{1 - \pi i}})} } \right]\frac{{1 - {P_{rk}}}}{{CSR - {N_v} + 1}}} \right]^{{N_v} - 1}}
				\end{array}	
			\end{aligned}  ,
			\label{collision}
		\end{equation}
		$\pi$ is the probability that the vehicle is at the moment of preparing to select resources, that is, the probability that all three conditions of the vehicle queue being non empty, RC being 0, and rebooking new resources are met simultaneously. $\Gamma$ is equivalent to the number of sub frames in SW, CSR represents the total number of resources in SW, which is the product of $\Gamma$ and the number of sub-channels. Therefore, CSR is proportional to $\Gamma$, and $N_v$ is the total number of vehicles. 
	
		So when $\Gamma$ remains constant, $P_{col}$ increases with $N_v$. The expression for the SINR of the receiver when a collision occurs is
		\begin{equation}
			\begin{aligned}
				\eta _{i \to j}^{t,n} = \frac{{p_i^t{{\left| {h_{i \to j}^{t,n}} \right|}^2}}}{{\sum\limits_{g \in {N_g}} {p_g^t{{\left| {h_{g \to j}^{t,n}} \right|}^2} + } p_n^2}}
			\end{aligned}  ,
		\end{equation}
		$g$ represents interfering vehicles, and $p^t_g$ is their transmitted energy. When the same resource is shared by multiple vehicles, the $\eta _{i \to j}^{t,n}$ drops, which can cause transmission failures if it's too low. Therefore, introducing SIC based NOMA to address this situation. The receiver sorts the received multiple signals according to the received power, and use the signal with the strongest received power as the target and treat other signals as interference, then decode and remove them. Then loop through the operation until the SINR of all signals is calculated. If ${N_k} = \left\{ {k \in N\backslash i|p_{i \to j}^t{{\left| {h_{i \to j}^{t,n}} \right|}^2} > p_{k \to j}^t{{\left| {h_{k \to j}^{t,n}} \right|}^2}} \right\}$ is a vehicle group with weaker receiving power than vehicle $i$, then the SINR at vehicle $j$ for vehicle $i$ is
		\begin{equation}
			\begin{aligned}
				\eta _{i \to j}^{t,n} = \frac{{p_i^t{{\left| {h_{i \to j}^{t,n}} \right|}^2}}}{{\sum\limits_{k \in {N_k}} {p_k^t{{\left| {h_{k \to j}^{t,n}} \right|}^2} + } p_n^2}}
			\end{aligned}  ,
		\end{equation}
		
		It can be seen that when collisions occur, 
		messages are transmitted over the same frequency band but with different power levels. By using NOMA to sequentially decode higher power messages first, the receiver can effectively cancel out the stronger messages before decoding the weaker ones. This process mitigates interference from stronger messages allowing the weaker signals to be decoded more accurately.
		Thereby increasing SINR and reducing the possibility of transmission failure caused by SINR below the threshold. And the transmission time between vehicle $i$ and $j$ is
		\begin{equation}
			\begin{aligned}
				l_{i \to j}^t = \frac{G}{{{W_i}{{\log }_2}(1 + \eta _{i \to j}^t)}}
			\end{aligned}  .
		\end{equation}

		\subsection{Energy consumption model}
		In C-V2X, the vehicle will communicate throughout the entire time slot of the reserved resources, so the energy consumption generated by the vehicle during each transmission is
		\begin{equation}
			\begin{aligned}
				\varepsilon _i^t = p_i^t\beta _i^t
			\end{aligned}  .
		\end{equation}
		
		However, the vehicle will repeatedly occupy $RC^0_i$ times after each reserved resource. 
		So $E_i^t$, the total energy consumption of the vehicle during this time period, is represented as
		\begin{equation}
			\begin{aligned}
				E_i^t = \varepsilon _i^tRC_i^0
			\end{aligned}  .
			\label{eq12}
		\end{equation}

		\subsection{Problem formulation}
		Establish a joint optimization problem with the objective of minimizing the weighted average of AoI and energy consumption for all vehicles. Because the transmission interval and power determine the magnitude of AoI and energy consumption, they are used as variables for the optimization problem. Therefore, the optimization objective is
		\begin{align}
			\label{eqmin}
			&\min_{{\Gamma}^t,{p}^t} \left[\omega_1\overline{E}+\omega_2\overline\Phi \right]\\
			s.t. \qquad &p^{t}\in [0,P_{max}], \forall t \in \mathcal{T}, \tag{\ref{eqmin}{a}} \label{eqmina}\\
			&\Gamma^{t}\in\{20,50,100\}, \forall t \in \mathcal{T}, \tag{\ref{eqmin}{b}} \label{eqminb}
		\end{align}
		among them, $\omega_1$ and $\omega_2$ are non negative weight factors, and $\mathcal{T}=\{1, \ldots, t, \ldots, T\}$ is the set notation of time slots. Where $\overline{E}$ and $\overline{\Phi}$ represent the average energy consumption and the average AoI at the receiving end, respectively, and their expressions are:
		\begin{equation}
			\begin{aligned}
				\bar E = \frac{1}{T}\frac{1}{{{N_v}}}\sum\limits_{t \in {\cal T}} {\sum\limits_{i \in {N_v}} {E_i^t} } 
			\end{aligned}  .
			\label{eq12}
		\end{equation}
		\begin{equation}
			\begin{aligned}
				\bar \Phi  = \frac{1}{T}\frac{1}{{{N_v}}}\frac{1}{{{N_v} - 1}}\sum\limits_{t \in {\cal T}} {\sum\limits_{i \in {N_v}} {\sum\limits_{j \in {N_v}} {\Phi _{i \to j}^t} } } 
			\end{aligned}  .
		\end{equation}

		\section{DRL Method for Optimization of RRI and Power Allocation}
		\label{DRL Method for Optimization of RRI and Power Allocation}
		Considering the uncertain channel conditions of the C-V2X system, this section introduces an Multi-Pass deep Q-Networks (MPDQN) method based on DRL to solve the $\Gamma$ and $p$ allocation problems of vehicles in the system. Using the DRL framework to model the problem, which mainly includes states, actions, policies, and rewards. Specifically, the RSU takes action $a^t$ based on the current state $s^t$ and the policy, and obtains the corresponding reward $r^t$ and the state $s^{t+1}$ for the next time slot. 
		
		\subsection{DRL framework}
		\begin{itemize}
			\item \textbf{State:} In time slot $t$, vehicle $i$ uses broadcasting to communicate with other vehicles. The number of receivers for vehicle $i$ is $N^t$, and the average distance between them and vehicle i is $d_i$. When vehicle $i$ transmits messages in time slot $t$, the ratio of the number of successful recipients to the total number is $Rn$, reflecting the impact of current transmission power on the communication process under uncertain channel conditions. $RC^0$ affects the queue message processing rate and communication frequency. Therefore, the state of vehicle $i$ in the time slot is defined as
			\begin{equation}
				{s}^t_i = [N^t_i,d^t_i,Rn^t_i,RC^0_i]
				.
				\label{state}
			\end{equation}
			
			\item \textbf{Action:} RSU allocates transmission intervals and power to vehicles based on their status, therefore the actions of vehicle $i$ in time slots are defined as
			\begin{equation}
				{a}^t_i = [{\Gamma}^t_i, {p}^t_i]
				.
				\label{action1}
			\end{equation}
			The $\Gamma$ is a discrete value and $p$ is a continuous value, so the MPDQN algorithm that can simultaneously make choices for actions in this situation is used to solve it. First, match each discrete action with a continuous action as part of the state, treating the two actions as one action tuple, and then select the action tuple based on the state. Therefore, in MPDQN, the action tuple of a vehicle is defined as
			\begin{equation}
				{a}^t_i = ({\Gamma}^t_i, {p}_\Gamma)
				.
				\label{action2}
			\end{equation}
			
			\item \textbf{Reward function:}
			In this chapter, the goal of RSU is to optimize communication performance in the system, including AoI and energy consumption. Every time a vehicle selects an action, it will use it until the next reserved resource is selected. Therefore, the energy consumption generated by vehicle $i$ during that time period and the AoI of its receiver will be weighted and averaged as the reward function. So the reward function for vehicles in time slot $t$ is defined as
			\begin{equation}
				{r}^t_i=-[\omega_{1}E^t_{i}+ \omega_2\overline{\Phi^t_{i}}]
				,
			\end{equation}
			$\overline{\Phi^t_{i}}$ is the average AoI received by vehicle i at the receiving end during this time period, expressed as
			\begin{equation}
				\overline {\Phi _i^t}  = \frac{1}{T}\frac{1}{{{N_v}}}\sum\limits_{t = 1}^T {\sum\limits_{j \in {N_v}}^{} {\Phi _{i \to j}^t} } 
				.
			\end{equation}
		\end{itemize}

		\subsection{Optimizing allocation based on MPDQN}
		As described in \cite{MPDQNNR}, the optimization objective is achieved by adjusting the values of discrete action $\Gamma$ and continuous action $p$. This means that the action space is not solely discrete or continuous, which makes traditional DRL methods inadequate. Therefore, MPDQN is needed to directly handle this situation. The agent using MPDQN first evaluates the value of each discrete action and corresponding continuous action based on the state and $\theta_x$. It then pairs each discrete action with the continuous action that has the highest corresponding value, forming an action tuple. Finally, it identifies the action tuple with the highest value based on the state and $\theta _Q$, determining the optimal discrete and continuous actions for the current state.
		
		According to \cite{MPDQN3,MPDQN4}, for a set of action $({\Gamma}^t_i, {p}_\Gamma)$, its state action value function can be expressed as $Q(s^t,a^t) = Q\left( {s^t,\left( {\Gamma ,{p_\Gamma }} \right)} \right)$.
		To correspond $\Gamma$ and $p$, define a policy network as the transition function between discrete and continuous actions in state $s^t$
		\begin{equation}
			\begin{aligned}
				p_\Gamma ^t = {x^Q}\left( {{s^t},{\Gamma}^t ;{\theta _x}} \right)
			\end{aligned}  ,
			\label{xQ}
		\end{equation}
		where $\theta_x$ is the weight of the network.
		
		Therefore, the action value function of vehicle $i$ in state $s^t_i$ and action $({\Gamma}^t_i, {p}_\Gamma)$ is
		\begin{equation}
			\begin{aligned}
				Q\left( {s_i^t,\left( {\Gamma _i^t,{p_\Gamma }} \right)} \right) = Q\left( {s_i^t,\left( {\Gamma _i^t,{x^Q}\left( {s_i^t,\Gamma ;{\theta _x}} \right)} \right)} \right)
			\end{aligned}  .
			\label{Qvaluex}
		\end{equation}
		
		Using a deep neural network with network weight $\theta_Q$ to approximate $Q(s,(\Gamma),p)$, the Eq. \ref{Qvaluex} can be written as
		\begin{equation}
			\begin{aligned}
				Q(s_i^t,a_i^t) = Q\left( {s_i^t,\left( {\Gamma _i^t,{x^Q}\left( {s_i^t,\Gamma ;{\theta _x}} \right)} \right);{\theta _Q}} \right)
			\end{aligned}  .
			\label{QvalueQ}
		\end{equation}
		
		Then, by updating the network weights, the optimization objective of the problem is approached, and the target value of the vehicles in the scene is defined as
		\begin{equation}
			\begin{aligned}
				y_{}^t = r_{}^t + \gamma \mathop {\max }\limits_\Gamma  Q\left( {{s^{t + 1}},{\Gamma ^{t + 1}},{x^Q}\left( {{s^{t + 1}};{\theta _x}} \right);{\theta _Q}} \right)
			\end{aligned}  .
		\end{equation}
		
		The loss function of each network is defined as
		\begin{equation}
			\begin{aligned}
				\begin{array}{l}
					L_{Q}\left(\theta_{Q}\right)= 
					\underset{\left(s^t, (\Gamma, p_{\Gamma}), r^t, s^{t+1}\right) \sim M}{\mathbb{E}}\left[\frac{1}{2}\left(y^{t}-Q\left(s^t,\left(\Gamma, p_{\Gamma}\right) ; \theta_{Q}\right)\right)^{2}\right]
				\end{array}
			\end{aligned}  ,
			\label{LossQ}
		\end{equation}
		\begin{equation}
			\begin{aligned}
				\begin{array}{l}
					L_{x}\left(\theta_{x}\right)= 
					\underset{s^t \sim M}{\mathbb{E}}\left[- \underset{{\Gamma=\{20,50,10\}}}\sum Q\left(s^t, \Gamma, x^{Q}\left(s, \Gamma ; \theta_{x}\right) ; \theta_{Q}\right)\right]
				\end{array}
			\end{aligned} .
			\label{Lossx} 
		\end{equation}

		Finally, update the parameters through the following methods
		\begin{equation}
			\begin{aligned}
				\theta _Q^{t + 1} = \theta _Q^t - l{r_Q}{\nabla _{{\theta _Q}}}{L_Q}\left( {{\theta _Q}} \right)
			\end{aligned}  ,
			\label{updateQ}
		\end{equation}
		\begin{equation}
			\begin{aligned}
				\theta _x^{t + 1} = \theta _x^t - l{r_x}{\nabla _{{\theta _x}}}{L_x}\left( {{\theta _x}} \right)
			\end{aligned}  ,
			\label{updatex}
		\end{equation}
		the $lr_x$ and $lr_Q$ are the learning rates of the two policy networks.

		Next, we will explain the algorithm in detail, with the pseudocode in Algorithm \ref{al1}. Firstly, randomly initialize $\theta_x$ and $\theta_Q$, and establish an experience replay buffer $M$. Next, the algorithm loops through $EP$ segments, resetting the system model simulation parameters at the beginning of each segment. The intelligent agent outputs the action tuple $({\Gamma}^t_i, {p}_\Gamma)$ based on the initial network parameters, and then observes the state ${s}^1_i = [N^1_i,d^1_i,Rn^1_i,RC^0_i]$ after the action is used. Then, the algorithm cycles from time slot 1 to time slot $T$. For each time slot $t$, it randomly selects with a certain probability or calculates the corresponding $p$ for each $\Gamma$ according to formula Eq. \ref{xQ}. Then, according to Eq. \ref{QvalueQ}, the action tuple $({\Gamma}^t_i, {p}_\Gamma)$ with the highest $Q$ value is obtained, and exploration noise is added. Finally, the intelligent agent will store the observed reward $r$, state $s^{t+1}$, and previous state actions as tuples  $[s^t,(\Gamma^t,p^t),r^t,s^{t+1}]$ in $M$. When the number of tuples exceeds the size of the sample $B$, $B$ tuples are taken from it to update the network parameters.
		\begin{algorithm}
			\caption{Optimization algorithm based on MPDQN}
			\label{al1}
			\KwIn{$\gamma$, $\tau_Q$, $\tau_x$, $\theta_Q$, $\theta_x$}
			\KwOut{optimized $\theta_Q$, $\theta_x$}
			Randomly initialize the $\theta_Q$, $\theta_x$\;
			Initialize replay experience buffer $M$\;
			\For{episode from $1$ to $EP$ }
			{
				Reset simulation parameters for the system model\;
				Receive initial observation state $\boldsymbol{s}_{0}$\;
				\For{slot $t$ from $1$ to $T$ }
				{
					Generate action tuples\;
					Execute action $\left( {\Gamma _i^t,{p_\Gamma }} \right)$, observe reward $r^t$ and new state $\boldsymbol{s}^{t+1}$ from the system model\;
					Store transition tuple $\left[ {{s^t},\left( {{\Gamma ^t},{p_\Gamma }} \right),{r^t},{s^{t + 1}}} \right]$\;
					\If {number of tuples in $M$ is larger than $B$ }
					{
						Randomly sample a mini-batch of $B$ transitions tuples from $M$\;
						Update the critic network by minimizing the loss function according to Eq. \eqref{LossQ} and Eq. \eqref{Lossx}\;
						Update the actor network according to Eq. \eqref{updateQ} and Eq. \eqref{updatex}.
					}
				}
			}
			
		\end{algorithm}

		\section{Simulation Results and Analysis}
		\label{Simulation Results and Analysis}
		\subsection{Parameter Settings}
		In this section, we conducted extensive simulation experiments on the scene in Fig. \ref{fig1}, using Python 3.6 and MATLAB 2023b as simulation tools. Simulation was added and modified based on \cite{cecchini2017ltev2vsim}. Assuming the length of the highway $D$ is $500m$, the number of vehicles $N_v$ is $20-50$, and the communication distance $w$ of the vehicles is $150m$. According to the C-V2X standard, the bandwidth is set as 10 MHz and QPSK modulation for propagation. The $T_C$ of CAM takes $100ms$, and the other three types of messages are $\lambda$ set to 0.0001. The $T_H$, $T_D$, $K_H$ and $K_D$ are set to $100ms$, $500ms$, 8 and 5 respectively. In the simulation, the learning rates of the two networks are $lr_x=10^{-4}$ and $lr_Q=5*10^{-4}$, respectively, and the update parameters $\tau_x$ and $\tau_Q$ of the networks are both 0.01. Used replay memory size of 2000, sample size $B=128$, discount factor $\gamma= 0.99$. In the process of selecting actions, greedy action strategies and exploration of action parameters with additive Ornstein Uhlenbeck noise are used [59]. The parameters used in the simulation are shown in Table \ref{tab2}.
		
		\begin{table}
			\caption{Values of the parameters in the experiments.}
			\label{tab2}
			\footnotesize
			\centering
			\begin{tabular}{|c|c|c|c|}
				\hline
				\textbf{Parameter} &\textbf{Value} &\textbf{Parameter} &\textbf{Value}\\
				\hline
				$N_v$ &$20\sim{50}$ &$L$ &10\\
				\hline
				$D$ &$50 0m$ &$w$ &$150 m$\\
				\hline
				$T_C$ &$100 ms$  &$\lambda$ &$0.0001$ \\
				\hline
				$K_H$ &$8$  & $K_D$ &$5$ \\
				\hline
				$Lr_Q$ &$5*10^{-4}$ &$lr_{x}$ &$10^{-4}$\\
				\hline
				$P_{max}$ &$23 dBm$ &$v_{min}$ &$60 km/h$\\
				\hline
				$B$ &$128$  &$v_{max}$ & $80 km/h$\\
				\hline
				$\tau_Q$ &$0.01$ &$\tau_x$ &$0.01$ \\
				\hline
				$M$ &$2000$ &$\gamma$ &0.99\\
				\hline
			\end{tabular}
			\vspace{-0.5cm}
		\end{table}
		
		\subsection{simulation Result}
		The existing work adopts genetic algorithms and random strategies as baseline algorithms for resource allocation. Therefore, we chose these two algorithms for comparison. The introduction of random strategy and genetic algorithm here is as follows
		\begin{itemize}
			\item Random policy: In C-V2X, vehicles randomly allocate power for each vehicle within the range $[0, P_{max}]$ based on the utilization of multiple priority queues and NOMA. Additionally, each vehicle's RRI is randomly assigned from the set ${20, 50, 100}$.
			\item GA-based policy: In each time slot, the RSU randomly generates a population vector based on the range of RRI and power values, as well as the population size, and uses the reward function as the fitness function. Each individual in the population represents the RRI and power allocation for all vehicle. The RSU employs a competition method to select the better individuals from the population vector based on the fitness of all individuals for crossover, generating offspring. These offspring undergo mutation operations and are then provided to the vehicles to obtain new fitness. Through multiple iterations of evolution, the crossover and mutation operations explore better actions, approaching the optimal RRI and power allocation.
		\end{itemize}
		
		\begin{figure}[h]
			\centering
			\subfigtopskip=2pt 
			\subfigbottomskip=2pt 
			\subfigure[Single queue]{
				\label{single_low}
				\includegraphics[scale=0.45]{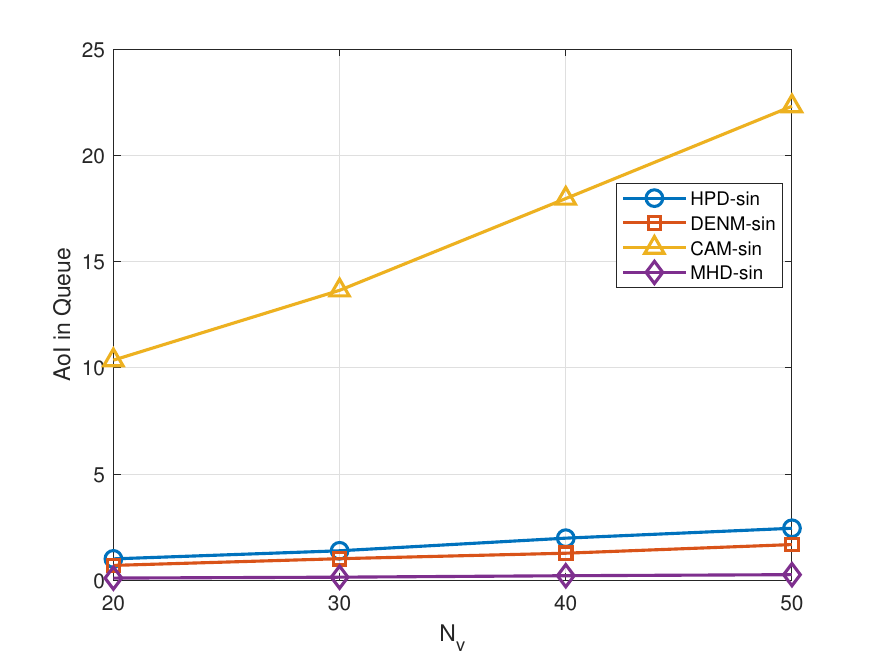}
			}
			\subfigure[Multi-priority queue]{
				\label{mul-low}
				\includegraphics[scale=0.45]{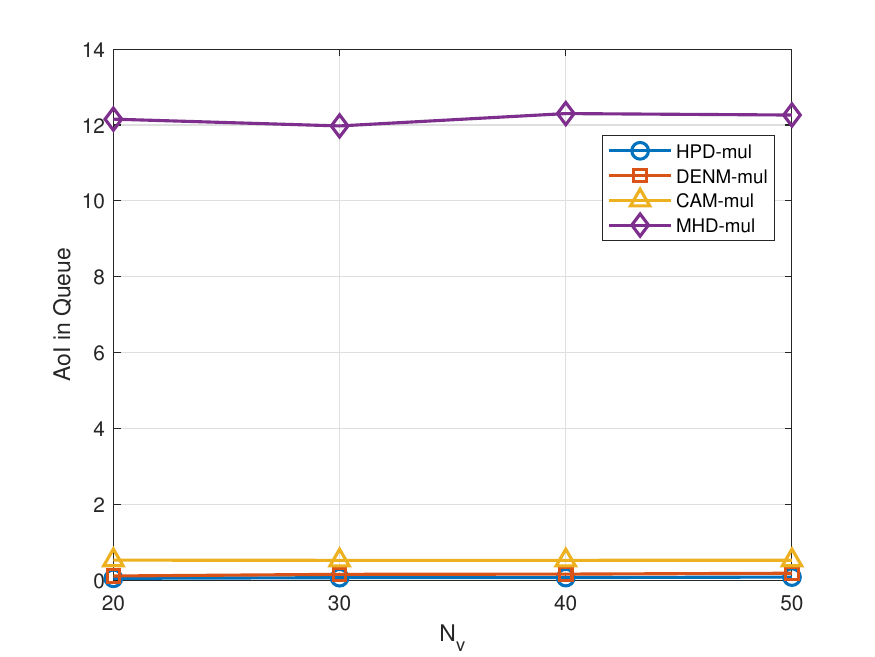}
			}
			\caption{The AoI in queue with low processing rates.}
			\vspace{-0.5cm}
		\end{figure}

		Fig. \ref{single_low} compares the average AoI variation for different types of messages from single-priority queue vehicles in the scenario when the $\Gamma$ is selected as $100ms$. It is evident that as the number of vehicles increases, the average AoI of each message type in the queue gradually increases. Moreover, since the processing rate of vehicle message queues is $10/s$ when the $\Gamma=100ms$, and the sum of production rates for all message types is significantly higher than this value, all message types struggle to be transmitted in a timely manner. Therefore, the relative sizes of AoI among these messages are correlated with their respective production rates. Messages with higher production rates exhibit larger AoI values. Consequently, the average AoI of the CAM type, which has the highest production rate, is the largest. While the average AoI of the MHD message type, which has the least frequent production rate, is the smallest.

		Fig. \ref{mul-low} compares the average AoI variations of messages of various types for multi-priority queue vehicles in the scenario when the RRI is 100ms. It can be observed that the higher the priority of a message, the smaller its average AoI, while the lower the priority, the larger the average AoI, ensuring the freshness of critical information when not all information transmission can be guaranteed in the communication environment. Moreover, compared to various types of messages in a single-priority queue, the average AoI of messages with higher priority becomes smaller. This is because in the presence of priorities, messages with lower priority always need to wait for messages with higher priority to be transmitted first, resulting in a lower processing rate for the queue of lower-priority messages compared to the queue of higher-priority messages. Among them, CAM-type messages show the largest variation because they have the highest message generation frequency, making them more prone to accumulation, especially when all types of messages are in a single queue. Therefore, when the priority of CAM-type messages is higher than that of MHD-type messages, the freshness of lower-priority MHD messages is sacrificed to ensure the transmission of CAM-type messages, thus reducing the average AoI.
		
		\begin{figure}[h]
			\centering
			\subfigtopskip=2pt 
			\subfigbottomskip=2pt 
			\subfigure[AoI in queue]{
				\label{queue}
				\includegraphics[scale=0.45]{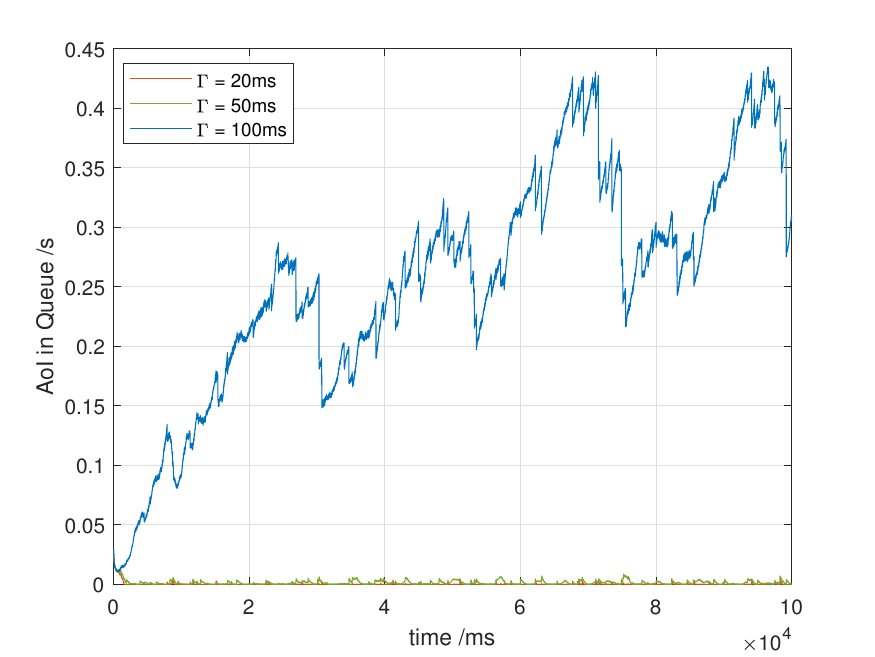}
			}
			\subfigure[AoI in receiver]{
				\label{receiver}
				\includegraphics[scale=0.45]{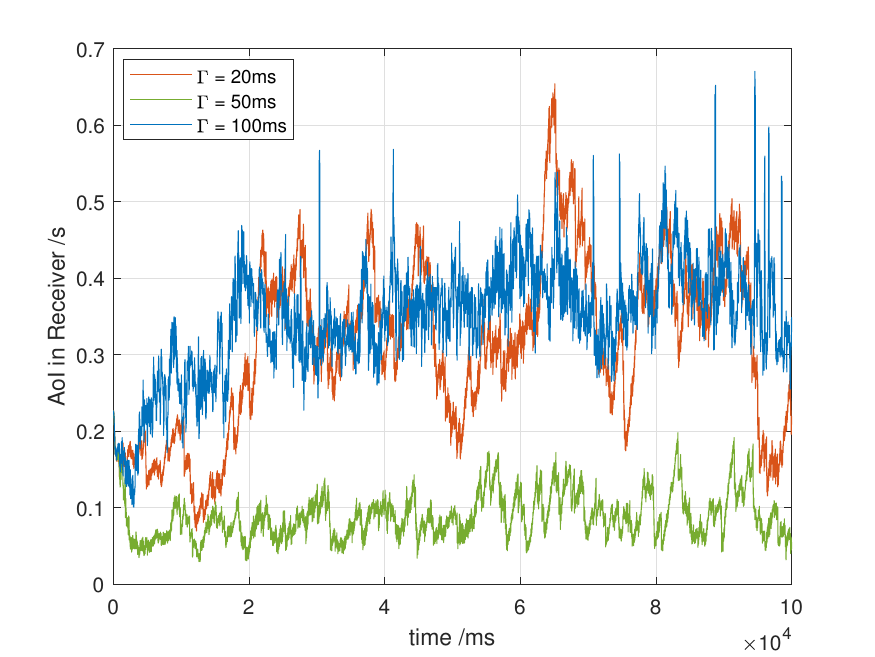}
			}
			\caption{The changes of average AoI with different $\Gamma$.}
			\vspace{-0.5cm}
		\end{figure}

		When $\Gamma$ is relatively small, the queue has a higher processing rate, resulting in smaller AoI for messages in the local queue. As shown in Fig. \ref{queue}, with $30$ vehicles in the scenario. Setting $\Gamma$ to $20$ or $50ms$ satisfies the transmission requirements of all queues, preventing message accumulation in vehicle queues and avoiding an increase in average AoI. However, when vehicles choose $\Gamma$ of $100ms$, it fails to meet the transmission demands of queue messages, especially occasional high-priority messages that further strain the processing rate, leading to an increase in AoI. Nevertheless, as depicted in Fig. \ref{receiver}, significant fluctuations occur in the AoI of receiver when the $\Gamma$ is at its minimum, while vehicles selecting an$\Gamma$ of $50ms$ consistently exhibit the lowest receiver AoI, significantly outperforming other selections. This indicates that vehicles choosing an  $\Gamma$ of $20ms$ are subjected to greater impact during transmission.

\begin{figure}[h]
			\centering
			\subfigtopskip=2pt 
			\subfigbottomskip=2pt 
			\subfigure[C-V2X without NOMA]{
				\label{CVran}
				\includegraphics[scale=0.45]{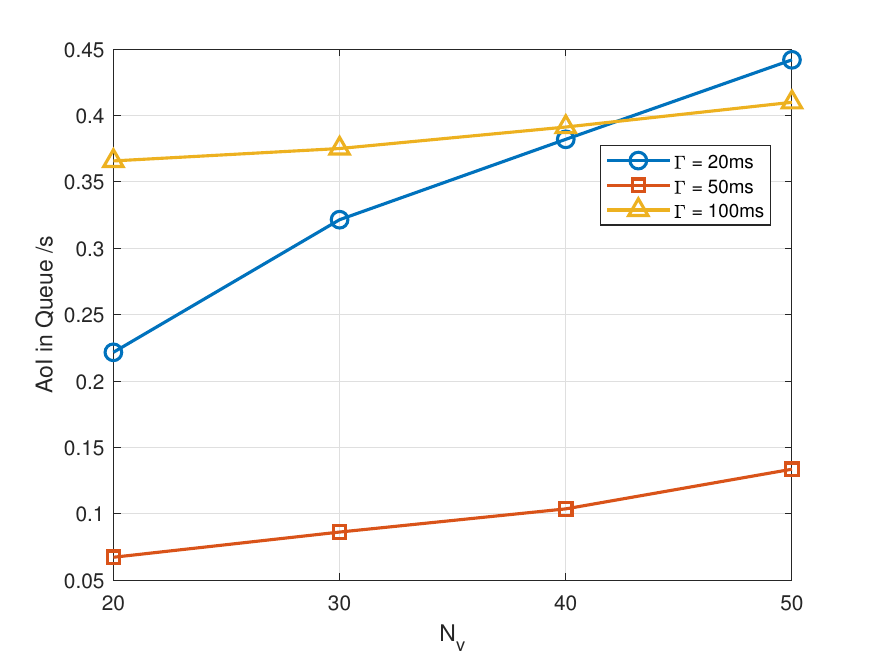}
			}
			\subfigure[C-V2X with NOMA]{
				\label{NOMA}
				\includegraphics[scale=0.45]{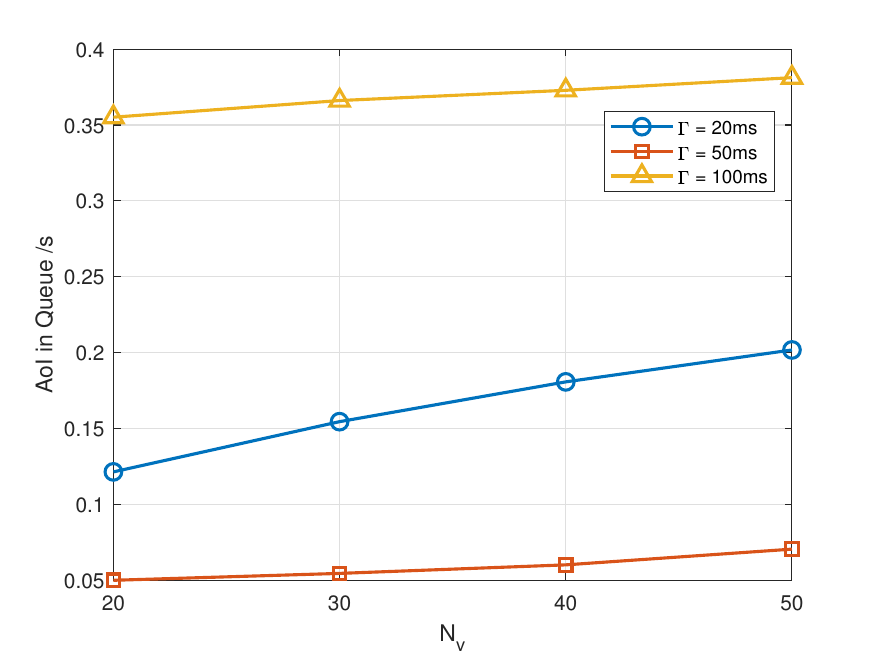}
			}
			\caption{The average AoI at receivers in C-V2X.}
			\vspace{-0.5cm}
		\end{figure}

		Fig. \ref{CVran} illustrates the influence of different RRI chosen by vehicles on the average AoI at the receiver for varying numbers of vehicles. It can be observed that the AoI is consistently higher when the $\Gamma$ is $100ms$, attributed to inadequate processing rates leading to excessively large queue AoI, coupled with longer communication intervals between the receiver and transmitter. Across all vehicle counts, the AoI of receiver is minimized when the $\Gamma$ is $50ms$, while in scenarios with fewer vehicles, the AoI with an $\Gamma$ of $20ms$ is smaller, but surpasses that of $100ms$ as vehicle count increases. Furthermore, regardless of the chosen $\Gamma$, the AoI of receiver increases as the vehicle count rises. This is due to an increase in collision probability during communication, as per Eq. (\ref{collision}). When the $\Gamma$ is $20ms$, the likelihood of contending for the same resources as other vehicles increases, leading to more instances of resource contention and thus higher chances of collisions. As the number of vehicles in the scenario increases, the required number of reserved resources also rises, increasing the likelihood of simultaneous resource selection and consequently elevating the possibility of different vehicles reserving the same resources.

		Fig. \ref{NOMA} illustrates the impact of vehicle selection of different RRI values on the average AoI at the receiver when using NOMA. Comparing with Fig. \ref{CVran}, it can be observed that NOMA improves the receiver AoI for all scenarios, particularly with the $20ms \ \Gamma$, where the improvement is most pronounced, reducing by approximately half. Moreover, as the $N_v$ increases, the improvement in each scenario becomes more significant, especially in situations with higher collision probabilities. Overall, NOMA can mitigate the effects of resource collisions resulting from the half-duplex selection of communication resources in C-V2X vehicular networking systems, thus optimizing the AoI of receiver.
		
		\begin{figure}[h]
			\centering
			\subfigtopskip=2pt
			\subfigbottomskip=2pt
			\subfigure[Reward curves]{
				\label{w1}
				\includegraphics[scale=0.45]{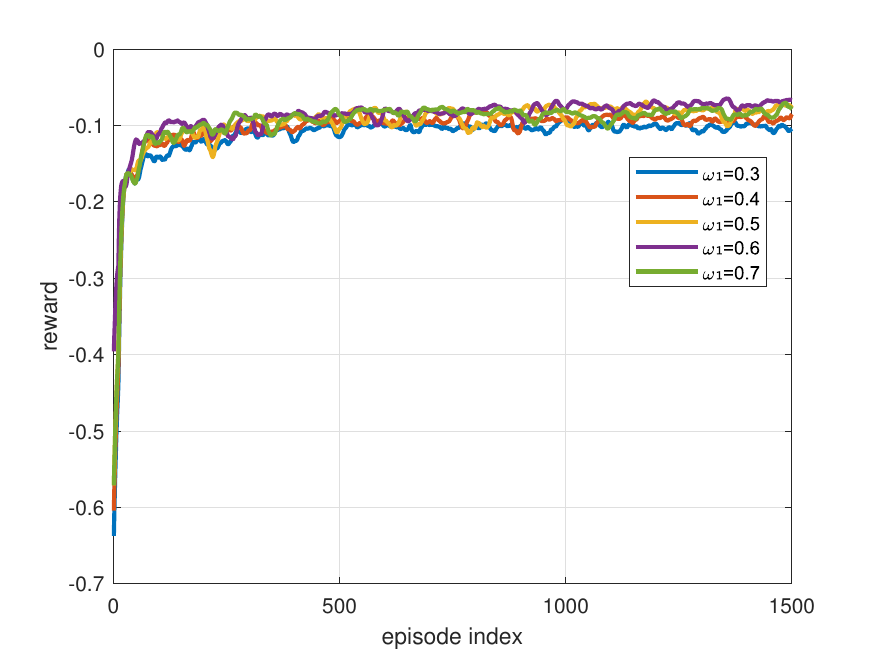}
			}
			\subfigure[Average reward]{
				\label{avgw1}
				\includegraphics[scale=0.45]{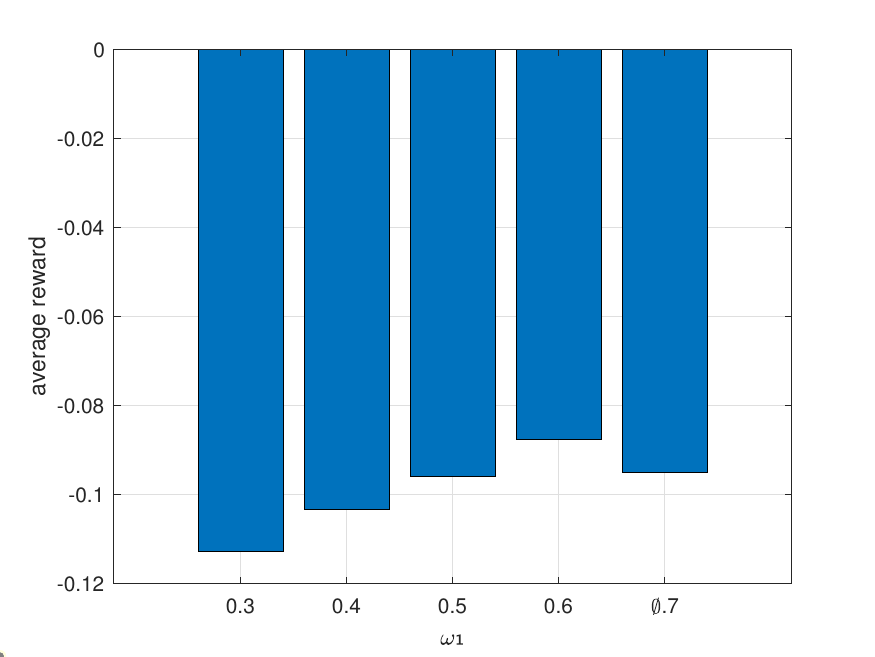}
			}
			\caption{The simulation results for different values of $\omega1$.}
			\vspace{-0.5cm}
		\end{figure}

		Fig. \ref{w1} illustrates the training process curves for different values of weights, ranging from $0.3$ to $0.7$. Values of $0.1$ and $0.2$ were excluded to prevent bias towards specific performance during learning. Each curve in the figure represents the average reward of all vehicles within a time slot in each segment. It can be observed that all curves fluctuate and rise moderately from $0$ to $500$ segments, as the agent searches for optimal RRI and power allocation parameters for the AoI of system and energy consumption. Subsequently, the curves fluctuate within a certain range, with significantly higher average rewards compared to the untrained state, indicating that the learned strategies of the agent are approaching optimality. However, the convergence fluctuation ranges vary for each curve when different network weight values are selected, especially evident in the significant difference between the best-performing curve at $0.6$ and the worst-performing one at $0.3$.

		To better compare the impact of different values of $\omega_1$ on rewards, the rewards of all vehicles across all segments were averaged, as shown in Fig. \ref{avgw1}. It can be observed that, consistent with Fig. \ref{w1}, although the average rewards are relatively high for different values of $\omega_1$, the highest average reward is achieved when $\omega_1=0.6$ compared to other scenarios. The largest difference is observed compared to $\omega_1=0.3$, nearing 0.02. Therefore, selecting 0.6 as the weight size for learning strategies in other scenarios seems justified.
		
		\begin{figure}[h]
			\centering
			\includegraphics[scale=0.45]{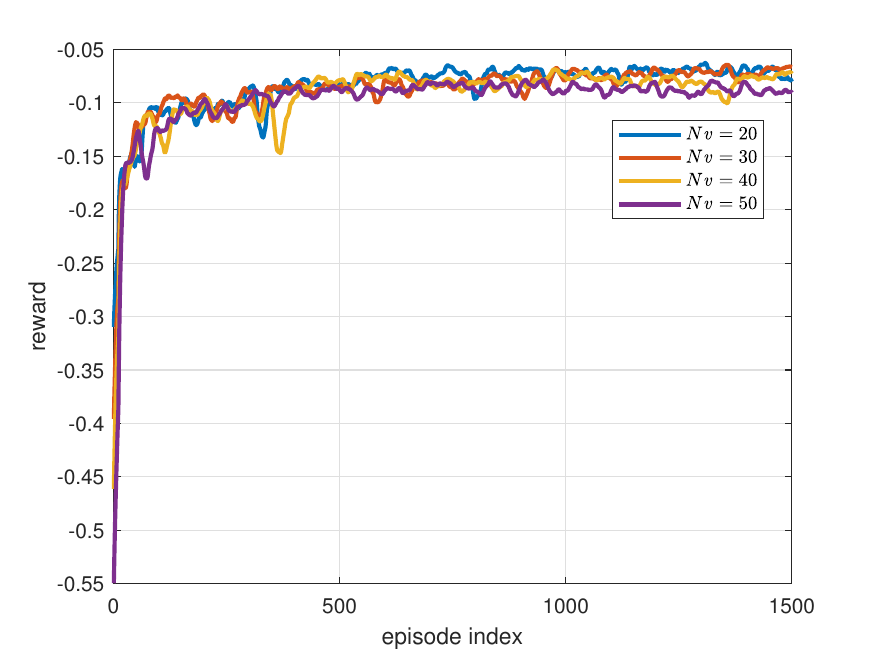}
			\caption{Reward curves for different $N_v$.}
			\label{reward}
			\vspace{-0.5cm}
		\end{figure}
		
		Fig. \ref{reward} displays how average rewards change with different $N_v$. Each curve in the figure represents the average reward of vehicles within each segment. It can be observed that all curves fluctuate upwards before segment 500 and stabilize within a certain range thereafter. By examining the fluctuation range of curves for different numbers of vehicles, it is evident that with fewer vehicles, the average reward fluctuates within a higher range. And the average reward gradually decreases as the $N_v$ increases. This is because the vehicle interference also rises, leading to an increase in AoI. To reduce the impact of increased AoI on rewards, the agent appropriately selects smaller  to increase the transmission frequency of vehicles, thereby reducing the AoI of receiver. However, smaller $\Gamma$ imply more transmissions, which may increase energy consumption and hence lower rewards.
		
		\begin{figure}[h]
			\centering
			\subfigtopskip=2pt 
			\subfigbottomskip=2pt 
			\subfigure[AoI]{
				\label{Nv-AoI}
				\includegraphics[scale=0.45]{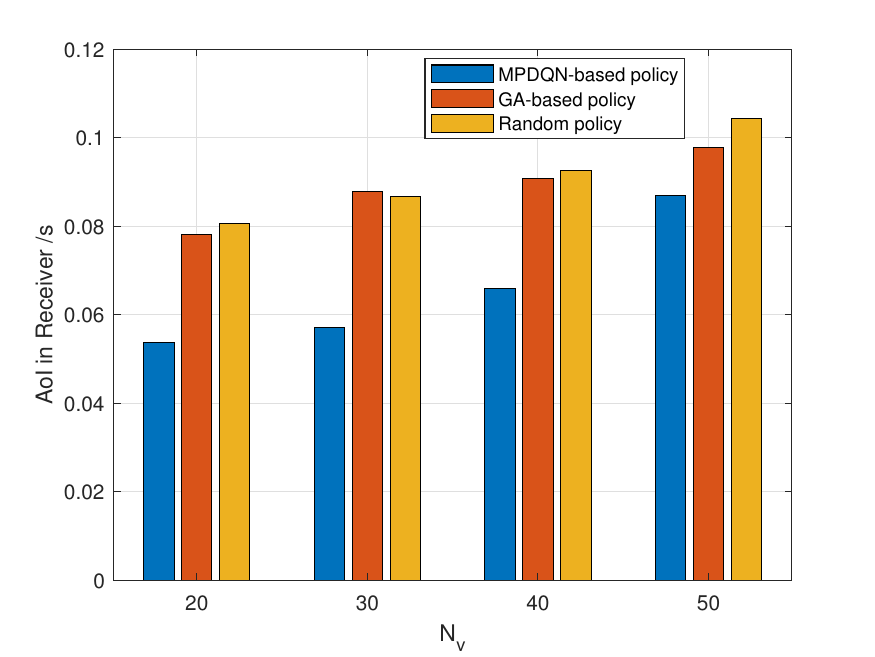}
			}
			\subfigure[Energy consumption]{
				\label{Nv-Energy}
				\includegraphics[scale=0.45]{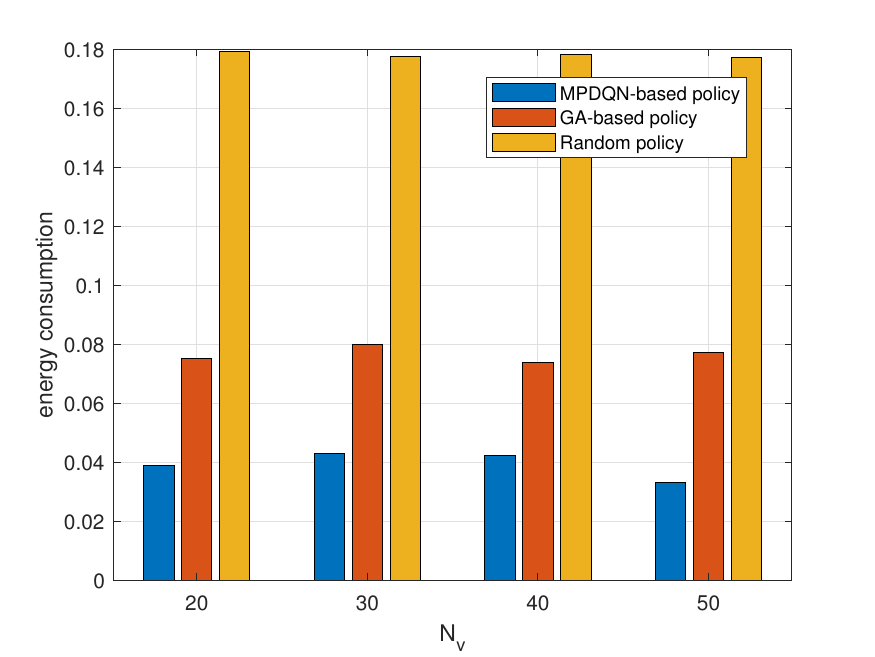}
			}
			\caption{The simulation results of each scheme for different numbers of vehicles.}
			\vspace{-0.5cm}
		\end{figure}
		
		Fig. \ref{Nv-AoI} compares the average sum AoI under MPDQN, GA, and random algorithms. It can be observed that under all three strategies, the average sum AoI within the system tends to increase as the $N_v$ increases. This is because the average AoI encompasses the AoI of all vehicles in the system acting as receivers, so as the $N_v$ increases, both the queuing time for messages and the transmission process affect more receivers. Particularly, interference generated during transmission increases with the $N_v$, leading to an increase in the average AoI. Additionally, it can be seen that both MPDQN and GA algorithms generally achieve lower average AoI compared to the random strategy, with MPDQN achieving the lowest average AoI. This is because MPDQN can adaptively adjust the RRI and power based on the current state of the vehicles.

		Fig. \ref{Nv-Energy} compares the average energy consumption within the system under MPDQN, GA algorithm, and random algorithm for different $N_v$. It can be seen that the energy consumption fluctuates slightly with an increase in the $N_v$ under MPDQN and GA algorithms, while the average energy consumption remains relatively stable with an increase in the seen under the random algorithm. This is because the average energy consumption is the mean of the energy consumption of all vehicles in the scenario. When using the random algorithm, the RRI and power of all vehicles in the scenario are randomly chosen, and an increase in the $N_v$ does not significantly affect the mean energy consumption of all vehicles.
		However, for MPDQN and GA algorithms, both AoI and energy consumption need to be considered simultaneously. As mentioned in Fig. \ref{Nv-Energy}, when the $N_v$ increases, the average AoI increases. However, since the weight of energy consumption is set larger than the weight of AoI, MPDQN and GA algorithms prioritize selecting relatively lower energy consumption while ensuring that the AoI remains within an acceptable range. Therefore, despite the increase in the number of vehicles, the variation in energy consumption is not as significant as the variation in AoI.
		
		\begin{figure}[h]
			\centering
			\subfigtopskip=2pt 
			\subfigbottomskip=2pt 
			\subfigure[AoI]{
				\label{size-AoI}
				\includegraphics[scale=0.45]{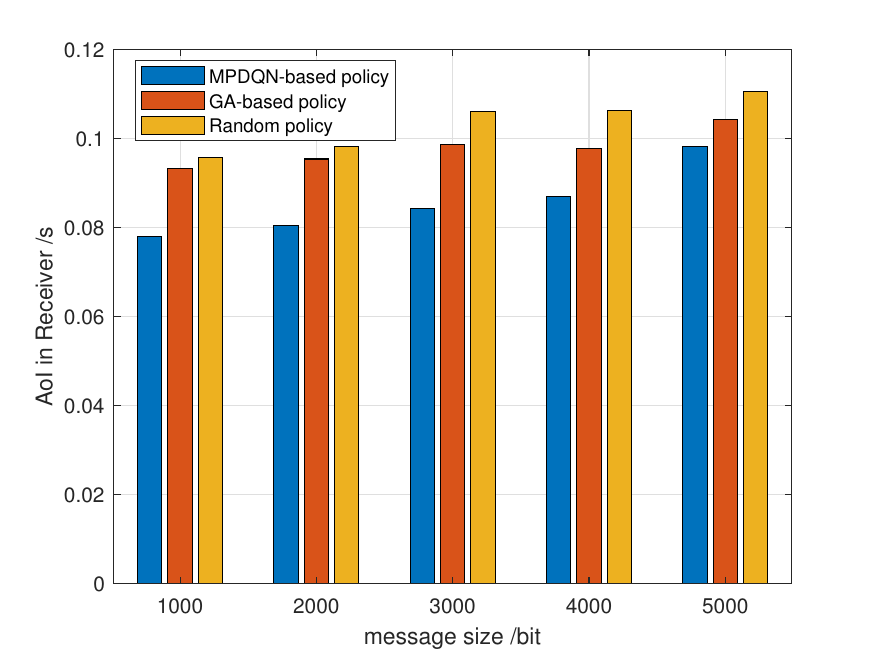}
			}
			\subfigure[Energy consumption]{
				\label{size-Energy}
				\includegraphics[scale=0.45]{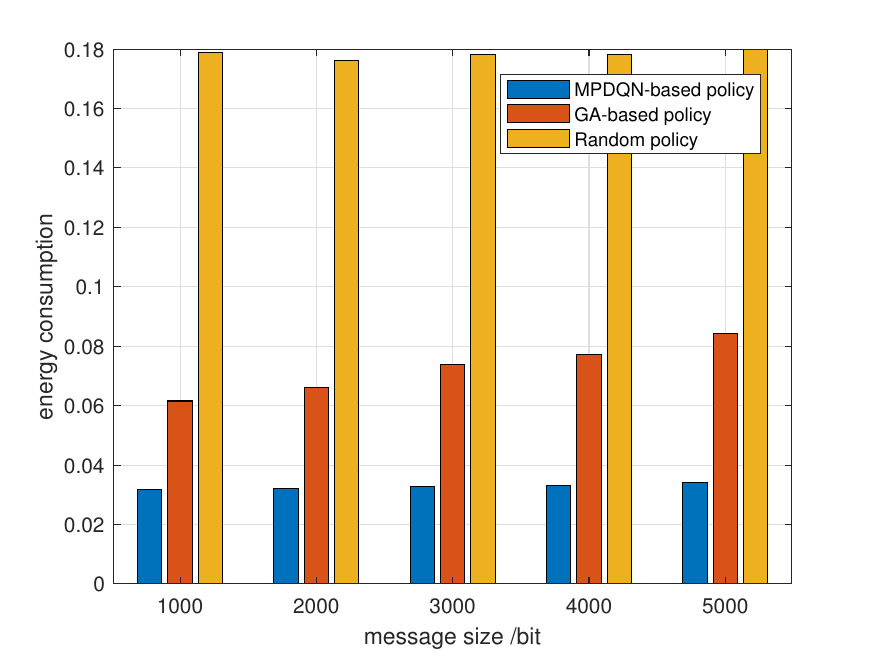}
			}
			\caption{The simulation results of each scheme for different packet size.}
			\vspace{-0.5cm}
		\end{figure}
		
		Fig. \ref{size-AoI} compares the average AoI under different message sizes for MPDQN, GA algorithm, and random algorithm when the number of vehicles is 50. It can be observed that the average AoI at the receiver rises with the message size. This is because larger data packets not only increase the transmission delay at a constant power but also may lead to transmission failure when the transmission delay exceeds the time slots of the communication resources. However, both MPDQN and GA algorithms can mitigate this impact to some extent. Moreover, due to the real-time adjustment of RRI and power based on vehicle status, MPDQN achieves the best optimization effect on the average AoI at the receiver.

		Fig. \ref{size-Energy} compares the system average energy consumption under different message sizes for MPDQN, GA algorithm, and random algorithm when the number of vehicles is 50. It can be observed that under the random algorithm, there is almost no difference in average energy consumption. This is because when messages become larger, vehicles only transmit within a short time frame of the reserved resources, resulting in minimal impact on the energy consumption per transmission. This also explains why there is little difference in energy consumption under different message sizes for the other two algorithms. However, due to the impact on AoI, energy consumption under the other two algorithms increases to ensure lower AoI. On the other hand, under different message sizes, the average aggregate energy consumption of MPDQN is always lower than that of the random algorithm and GA algorithm. This is because MPDQN can adaptively select RRI and power for vehicles based on their current status, and by determining the resource occupancy frequency through RRI, it minimizes the average energy consumption within the system.

		\section{Conclusions}
		\label{Conclusions}
		This paper mainly focuses on the C-V2X vehicular network system, multiple message types, NOMA, DRL, and communication resource allocation issues to optimize the average AoI and transmission energy consumption. Firstly, for the case of multiple signal types in C-V2X, a multi-priority queue is adopted to reduce the AoI of high-priority messages in the queue, and NOMA technology is introduced to reduce the impact of communication processes on the AoI. Secondly, in the C-V2X scenario, the MPDQN algorithm is used to process vehicle status information, determine the optimal resource allocation strategy, and select RRI and power for the vehicles to minimize the AoI and energy consumption of the entire system. Simulation results show that our method not only maintains the stability of vehicle communication but also ensures the timeliness of critical information, and outperforms baseline algorithms in optimizing average AoI and energy consumption.

\bibliographystyle{IEEEtran}
\bibliography{IEEEabrv,mybibfile}
	
\begin{IEEEbiography}[{\includegraphics[width=1in,height=1.25in,clip,keepaspectratio]{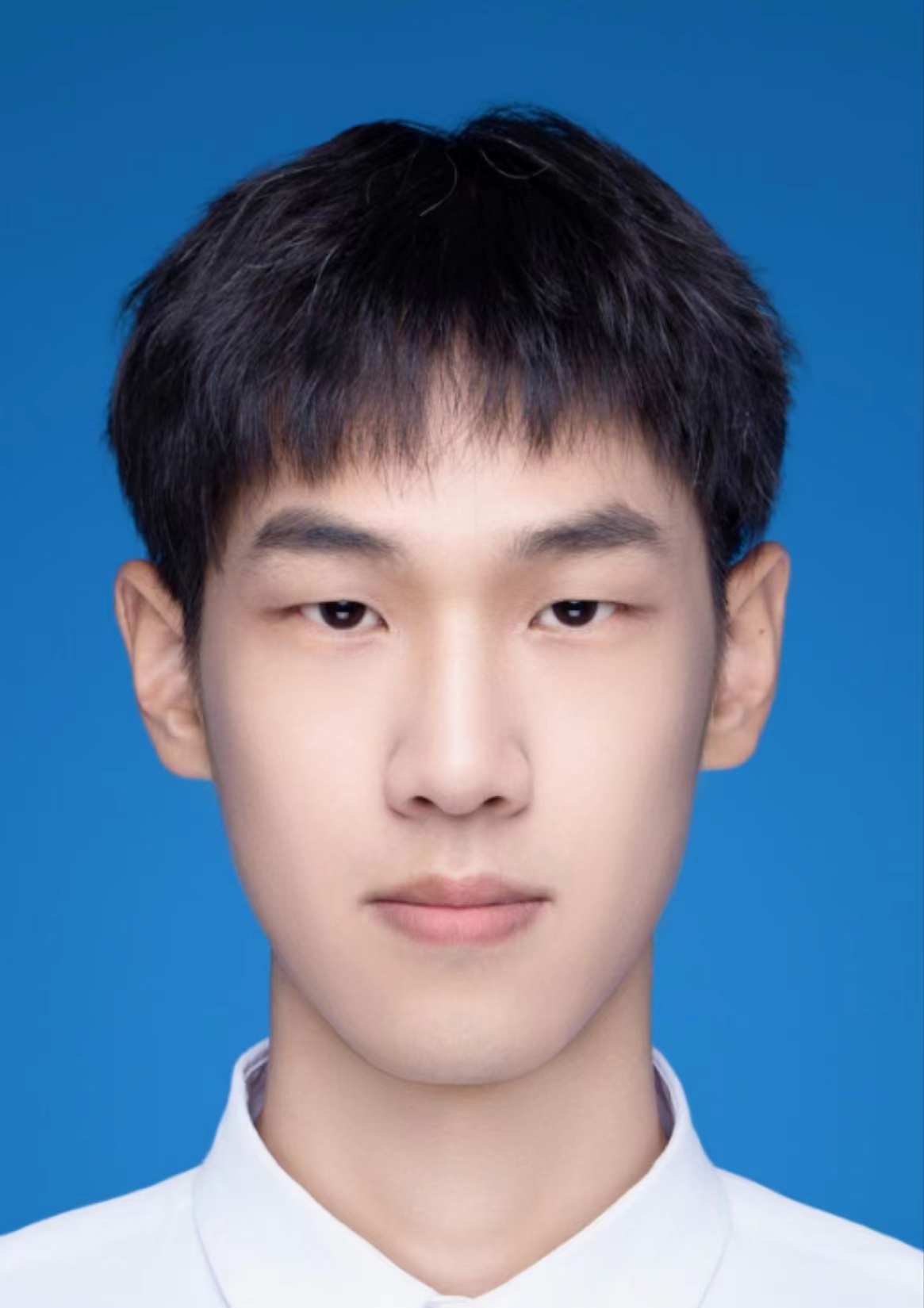}}] {Zheng Zhang} received the B.S. degree from the Huaiyin Institute of Technology, Huai'an, China, in 2021, and the M.S. degree from the Jiangnan University, Wuxi, China, in 2024. His research interests include deep reinforcement learning, C-V2X, age of information, and non-orthogonal multiple access.
\end{IEEEbiography}	

\begin{IEEEbiography}[{\includegraphics[width=1in,height=1.25in,clip,keepaspectratio]{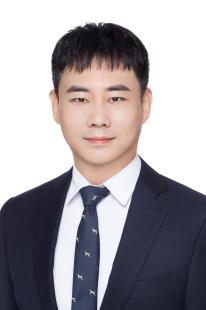}}] 
{Qiong Wu} (Senior Member, IEEE) received the Ph.D. degree in information and communication engineering from National Mobile Communications Research Laboratory, Southeast University, Nanjing, China, in 2016. From 2018 to 2020, he was a postdoctoral researcher with the Department of Electronic Engineering, Tsinghua University, Beijing, China. He is currently an associate professor with the School of Internet of Things Engineering, Jiangnan University, Wuxi, China.

Dr. Wu is a Senior Member of IEEE and China Institute of Communications. He has published over 80 papers in high impact journals and conferences, and authorized over 20 patents. He was elected as one of the world's top 2\% scientists in 2024 and 2022 by Stanford University. He has received the young scientist award for ICCCS'24 and ICITE’24. He has been awarded the National Academy of Artifical Intelligence (NAAI) Certified AI Senior Engineer. He was the excellent reviewer for Computer Networks in 2024. He has severed as the editorial board member of Sensors and CMC-Computers Materials \& Continua, the early career editorial board member of Radio Engineering and Chinese Journal on Internet of Things, the lead guest editor of Sensors, CMC-Computers Materials \& Continua and Frontiers in Space Technologies, the guest editor of Electronics and Chinese Journal on Internet of Things, the TPC co-chair of WCSP'22, the workshop chair of NCIC'23, ICFEICT'24, CIoTSC’24 and IAIC’24, as well as the TPC member and session chair for over 10 international Conferences. His current research interest focuses on vehicular networks, autonomous driving communication technology, and machine learning.
\end{IEEEbiography}


\begin{IEEEbiography}[{\includegraphics[width=1in,height=1.25in,clip,keepaspectratio]{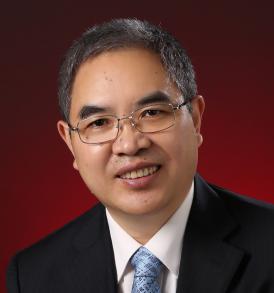}}] {Pingyi Fan} (Senior Member, IEEE) received the B.S. degree from the Department of Mathematics, Hebei University, in 1985, the M.S. degree from the Department of Mathematics, Nankai University, in 1990, and the Ph.D. degree from the Department of Electronic Engineering, Tsinghua University, Beijing, China, in 1994. From August 1997 to March 1998, he visited The Hong Kong University of Science and Technology as a Research Associate. From May 1998 to October 1999, he visited the University of Delaware, Newark, DE, USA, as a Research Fellow. In March. 2005, he visited NICT, Japan, as a Visiting Professor. From June 2005 to May 2014, he visited The Hong Kong University of Science and Technology for many times. From July 2011 to September 2011, he was a Visiting Professor at the Institute of Network Coding, The Chinese University of Hong Kong. He is currently a Professor and the director of open source data recognition innovation center at the Department of Electrical Engineering (EE), Tsinghua University. His main research interests include B5G technology in wireless communications, such as MIMO, OFDMA, network coding, network information theory, machine learning, and big data analysis.
Dr. Fan is a Member of The United States National Academy of Artifical Intelligence (NAAI), a Fellow of IET and an Overseas Member of IEICE. He is also a reviewer of more than 40 international journals, including 30 IEEE journals and eight EURASIP journals. He has received some academic awards, including the IEEE WCNC’08 Best Paper Award, the IEEE ComSoc Excellent Editor Award for IEEE TRANSACTIONS ON WIRELESS COMMUNICATIONS in 2009, the ACM IWCMC’10 Best Paper Award, the IEEE Globecom’14 Best Paper Award, the IEEE ICC’20 Best Paper Award, the IEEE TAOS Technical Committee’20 Best Paper Award,  IEEE ICCCS Best Paper Awards in 2023 and 2024, the CIEIT Best Paper Awards in 2018 and 2019, the DCASE Challenges Judge’s Award in 2024. He has attended to organize many international conferences, as the general chairs/TPC chair/ Plenary/Keynote speaker for over 30 international Conferences, including as the General Co-Chair of EAI Chinacom2020, 2023 and IEEE VTS HMWC 2014, the TPC Co-Chair of IEEE ICCCS2024 and a TPC Member of IEEE ICC, Globecom, WCNC, VTC, and Inforcom. He has served as an Editor for IEEE TRANSACTIONS ON WIRELESS COMMUNICATIONS, International Journal of Ad Hoc and Ubiquitous Computing (Inderscience), Journal of Wireless Communication and Mobile Computing (Wiley), Electronics (MDPI), and Open Journal of Mathematical Sciences, IAES international journal of artificial intelligence. 
\end{IEEEbiography}


\begin{IEEEbiography}[{\includegraphics[width=1in,height=1.25in,clip,keepaspectratio]{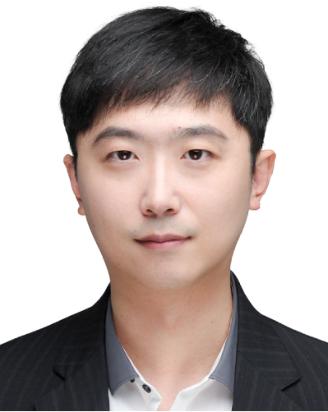}}] 
{Nan Cheng} received the B.E. and M.S. degrees from the College of Electronics and Information Engineering, Tongji University, Shanghai, China, in 2009 and 2012, respectively, and the Ph.D. degree from the Department of Electrical and Computer Engineering, University of Waterloo, Waterloo, ON, Canada, in 2016. He was a Postdoctoral Fellow with the Department of Electrical and Computer Engineering, University of Toronto, Toronto, ON, Canada, from 2017 to 2019. He is currently a Professor with the State Key Laboratory of ISN and with the School of Telecommunications Engineering, Xidian University, Xi’an, Shaanxi, China. He has published over 90 journal papers in IEEE Transactions and other top journals. His current research focuses on B5G/6G, AI-driven future networks, and space–air-ground integrated networks. Prof. Cheng serves as an Associate Editor for IEEE Transactions on Vehicle Technology, IEEE Open Journal of the Communication Society, and Peer-to-Peer Networking and Applications, and serves/served as a guest editor for several journals.
\end{IEEEbiography}

\begin{IEEEbiography}[{\includegraphics[width=1in,height=1.25in,clip,keepaspectratio]{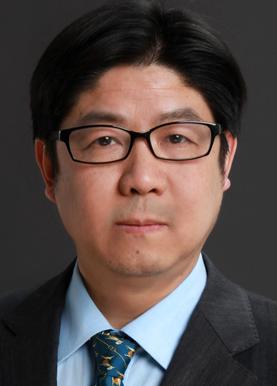}}] 
{Wen Chen} (M’03–SM’11) received the B.S. and M.S. from Wuhan University, China in 1990 and 1993 respectively, and PhD from University of Electro-communications, Japan in 1999. He is now a tenured Professor with the Department of Electronic Engineering, Shanghai Jiao Tong University, China, where he is the director of Broadband Access Network Laboratory. He is a fellow of Chinese Institute of Electronics and the distinguished lecturers of IEEE Communications Society and IEEE Vehicular Technology Society. He is the Shanghai Chapter Chair of IEEE Vehicular Technology Society, a vice president of Shanghai Institute of Electronics, Editors of IEEE Transactions on Wireless Communications, IEEE Transactions on Communications, IEEE Access and IEEE Open Journal of Vehicular Technology. His research interests include multiple access, wireless AI and RIS communications. He has published more than 200 papers in IEEE journals with citations more than11,000 in Google scholar.
\end{IEEEbiography}

\begin{IEEEbiography}[{\includegraphics[width=1in,height=1.25in,clip,keepaspectratio]{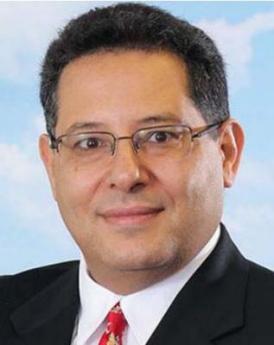}}] {Khaled Ben Letaief} (Fellow, IEEE) received the B.S. (Hons.), M.S., and Ph.D. degrees in electrical engineering from Purdue University, West Lafayette, IN, USA, in December 1984, August 1986, and May 1990, respectively. 
	
From 1990 to 1993, he was a Faculty Member with The University of Melbourne, Melbourne, VIC, Australia. Since 1993, he has been with The Hong Kong University of Science and Technology (HKUST), Hong Kong, where he is currently the New Bright Professor of Engineering. At HKUST, he has held many administrative positions, including an Acting Provost, the Dean of Engineering, the Head of the Electronic and Computer Engineering Department, and the Director of the Hong Kong Telecom Institute of Information Technology. He is an internationally recognized leader in wireless communications and networks. His research interests include artificial intelligence, mobile cloud and edge computing, tactile Internet, and sixth-genereation (6G) systems. In these areas, he has over 720 articles with over 44,450 citations and an H-index of over 100 along with 15 patents, including 11 U.S. inventions.
	
Dr. Letaief served as a member for the IEEE Board of Directors from 2022 to 2023. He is a member of the National Academy of Engineering, USA, and the Hong Kong Academy of Engineering Sciences; and a Fellow of the Hong Kong Institution of Engineers. He is well recognized for his dedicated service to professional societies and IEEE, where he served in many leadership positions, including the President of the IEEE Communications Society from 2018 to 2019, the world’s leading organization for communications professionals with headquarter in New York City, and members in 162 countries. He is recognized by Thomson Reuters as an ISI Highly Cited Researcher and was listed among the 2020 top 30 of AI 2000 Internet of Things Most Influential Scholars. He was a recipient of many distinguished awards and honors, including the 2007 IEEE Communications Society Joseph LoCicero Publications Exemplary Award, the 2009 IEEE Marconi Prize Award in Wireless Communications, the 2010 Purdue University Outstanding Electrical and Computer Engineer Award, the 2011 IEEE Communications Society Harold Sobol Award, the 2016 IEEE Marconi Prize Paper Award in Wireless Communications, the 2016 IEEE Signal Processing Society Young Author Best Paper Award, the 2018 IEEE Signal Processing Society Young Author Best Paper Award, the 2019 IEEE Communication Society and Information Theory Society Joint Paper Award, the 2021 IEEE Communications Society Best Survey Paper Award, and the 2022 IEEE Communications Society Edwin Howard Armstrong Achievement Award. He is the Founding Editor-in-Chief of the prestigious IEEE TRANSACTIONS ON WIRELESS COMMUNICATIONS. He has been involved in organizing many flagship international conferences.
\end{IEEEbiography}
\end{document}